\documentclass[11pt]{article}

\usepackage[margin=1in]{geometry}
\usepackage{setspace}
\usepackage[T1]{fontenc}
\usepackage[utf8]{inputenc}
\usepackage{lmodern}
\usepackage{float}

\usepackage{amsmath,amssymb,amsfonts}

\usepackage{graphicx}
\usepackage{subcaption}
\usepackage{tabularx}
\usepackage{colortbl}
\usepackage{xcolor}

\makeatletter
\newcommand{\thickcline}[1]{%
  \@thickcline #1\@nil%
}
\def\@thickcline#1-#2\@nil{%
  \omit
  \@multicnt#1%
  \advance\@multispan\m@ne
  \ifnum\@multicnt=\@ne\@firstofone{&\omit}\fi
  \@multicnt#2%
  \advance\@multicnt-#1%
  \advance\@multispan\@ne
  \leaders\hrule\@height3pt\hfill % thicker rule
  \cr
  \noalign{\vskip-1.5pt}%
}
\makeatother

\usepackage{microtype}
\usepackage{url}
\usepackage[colorlinks=true,linkcolor=blue,citecolor=blue,urlcolor=blue]{hyperref}
\usepackage{lineno} % optional

\usepackage[round,authoryear]{natbib}
\usepackage{authblk}

\title{SpikeDecoder: Realizing the GPT Architecture with Spiking Neural Networks}

\author[1,*]{Claas Beger}
\author[1]{Florian Walter}
\author[1]{Alois Knoll}

\affil[1]{Chair of Robotics, Artificial Intelligence and Real-time Systems\newline Technical University of Munich, Germany}

\affil[*]{Corresponding author: \href{mailto:claas.beger@tum.de}{claas.beger@tum.de}}

\date{}

\begin{document}
\maketitle

\begin{abstract}
The Transformer architecture is widely regarded as the most powerful tool for natural language processing, but due to a high number of complex operations, it inherently faces the issue of high energy consumption. To address this issue, we consider Spiking Neural Networks (SNNs), which are an energy-efficient alternative to conventional Artificial Neural Networks (ANNs) due to their naturally event-driven approach to processing information. However, this inherently makes them difficult to train. Often, many SNN-based models circumvent this issue by converting pre-trained ANNs. More recently, attempts have been made to design directly trainable SNN-based adaptations of the Transformer model structure. Although the results showed great promise, the application field was computer vision. Moreover, the proposed model incorporates only encoder blocks. In this paper, we propose SpikeDecoder, a fully SNN-based implementation of the Transformer decoder block, for applications in natural language processing. In a series of experiments, we analyze the impact of exchanging different blocks of the ANN model with spike-based alternatives to identify trade-offs and significant sources of performance loss. We further investigate the role of residual connections and the selection of SNN-compatible normalization techniques. Besides the work on the model architecture, we formulate and compare different embedding methods to project text data into spikes. Finally, we demonstrate that our proposed SNN-based decoder block reduces the theoretical energy consumption by 87\% to 93\% compared to the ANN baseline.
\end{abstract}


\section{Introduction}
Transformer models demonstrate notable proficiency in addressing tasks related to natural language, performing well in both sequence-to-sequence translation and language generation. More specifically, OpenAI's GPT models have garnered considerable public attention due to their impressive capabilities. However, their implementation, which is based on Artificial Neural Networks (ANNs), struggles with high energy consumption due to the complex computations required during inference. Spiking Neural Networks (SNNs) have experienced rising popularity as the third generation of neural networks (\cite{maass1997networks}, \cite{ghosh2009third}). Due to the all-or-nothing principle employed in their brain-inspired approach to information propagation, they have the potential to reduce energy consumption, particularly when executed on neuromorphic processors~\cite{walter2015nmc}. This could enable the efficient implementation of GPT-based models while preserving as much of the original performance as possible.

To design such a model, the self-attention mechanism proposed by \cite{vaswani2017attention} is a key work with high relevance. Their innovative approach has become the dominant one for natural language processing. However, while powerful, multi-head self-attention consists of a set of (at least) two matrix dot products and a softmax operation, which are by nature ill-suited for an SNN-based implementation.
This issue was addressed by \cite{zhou2022spikformer}, who introduced the Spiking Self Attention (SSA) block in an encoder-only architecture for image classification. While the model also includes common linear layers, its main components remain neuromorphic due to an alternating structure featuring spiking neurons. The Spikformer model outperformed the majority of SNN-based image classification models and even reached a competitive level with ANNs on some datasets. As only binary values are used, the matrix dot product can be formulated in a way that avoids MAC operations, and, in addition, the softmax operation also becomes obsolete. \cite{zhou2023spikingformer} extended the Spikformer model by adjusting the residual connections in order to prevent MAC operations on the sum of binary outputs passed to the linear layers. This is done by re-ordering the affected blocks and placing a spiking node directly after the summation operator. This not only reduces the cost of operations but also increases the model performance. Another SNN-based model that adapts the Vision Transformer architecture was proposed by \cite{li2022spikeformer}. They employ an adapted attention mechanism that considers both spatial and temporal information. However, this approach is not exclusively spiking and computationally more expensive than Spikformer. 
Apart from the decoder-based architecture, one of the stronger directly trained SNN networks is spike-element-wise ResNet, which was introduced in \cite{NEURIPS2021_afe43465} and outperformed all directly trained SNN models on major image datasets prior to Spikformer.

In the field of natural language processing, there have been very few attempts at SNN-based architectures. Most relevant is the work by \cite{zhu2023spikegpt}, who proposed SpikeGPT, the first spiking model directly trained for language generation. It is worth noting that this model does not use self-attention but instead employs RWKV, a technique based on Recurrent Neural Networks. Additionally, this model is not fully spiking and applies MAC operations at various points throughout the model. \cite{9664146} presented an effective way of ANN-SNN conversion, which adapts the structure and weights of a pre-trained Transformer model into a rate-coded SNN architecture. While the resulting model is fully spiking and performs well, it is not trained directly. Thus, a fully trainable spiking (decoder-only) Transformer language model has yet to be developed.

In this work, we present a novel, fully spiking model based on the GPT architecture for natural language generation. Additionally, we examine the effect of normalization, embedding, and residual connections on the resulting structure. Since there is no prior work on the development of a directly trainable SNN-based GPT model, we closely analyze the process of transitioning from an established ANN-based GPT architecture to a purely SNN-based model. Finally, we provide an outlook on optimization opportunities, both in terms of performance and pure spiking behavior. Since there is no prior benchmark for a comparable SNN architecture in this application field, we do not investigate performance optimizations. Instead, we focus on building a stable starting point. The proposed model structure features a novel decoder block and a spike-compatible token embedding. Moreover, we show how normalization, residual connections, and time step unification can be realized in the proposed architecture. We develop and evaluate our model to process single characters rather than than full words. This decision is primarily influenced by the complex task of translating natural language tokens into spikes while preserving information, which falls outside the scope of this work.

\section{Materials and Methods}
In the following subsections, we introduce the technical foundations and design choices that guide the development of our final SpikeDecoder model in Section~\ref{Results}. We first describe the spiking neuron model used throughout the architecture, including the Multi-Step Leaky Integrate-and-Fire dynamics and their implications for spike-based computation. We then discuss how textual inputs can be represented in a spike-compatible form, covering character embeddings, positional encodings, and the trade-offs between one-hot, binary, static, and learned representations. Next, we examine normalization strategies and their suitability for Transformer-like SNN architectures, before addressing the adaptation of residual connections to reduce floating-point operations while preserving trainability. We further consider how linear and normalization layers may be combined for deployment to improve computational efficiency. Finally, we describe strategies for handling the synthetic time-step dimension introduced by spiking computation and compare different approaches for reuniting temporal outputs before classification.

\subsection{Spiking Neuron Implementation}
\label{Choice of Spiking Neuron Implementation}
While several software libraries for SNNs are available, SpikingJelly \citep{SpikingJelly} is one of the most efficient \citep{Lenz_2023} and has been utilized in the research most relevant to this work, including Spikformer and SpikeGPT. Our model employs the Multi-Step Leaky Integrate-and-Fire (LIF) neuron. On the subthreshold level, the corresponding neuron dynamics are described in the following manner:
\begin{equation}\label{eq:2.0}
	V[t] = V[t-1] + \frac{1}{\tau}(X[t] - (V[t-1] - V_{reset}))
\end{equation}
$\tau$ is the membrane time constant, $V[t]$ is the voltage level at the respective time step $t$, and $X[t]$ is the input at time step $t$.
The dynamics of this model can be divided into three main phases: First, the input adds or subtracts a computed voltage according to the given membrane time constant and the potential from the previous time step. 
Second, a Heaviside step function is applied to the difference between the updated voltage and the spiking threshold, which corresponds to the neuron firing a spike when it exceeds the threshold. For backpropagation during training, this function is replaced by a differentiable surrogate. For our proposed model, we employ the Sigmoid function. Utilizing a surrogate gradient is a common practice that allows for mimicking spiking functionality in a way that enables gradient descent \citep{neftci2019surrogate}. Third, a neuronal reset is triggered immediately after a spike, causing the voltage to return to its reset value. By default, this value is set to zero.

For the Multi-Step LIF implementation, the input is expected to be a tensor containing a time step along the first dimension. It is processed by generating a spike flag and an internal voltage value for each timestep. 
In addition to integrating the incoming voltages, the model also features a leaky behavior, meaning that it loses voltage over time. This is realized by propagating a default value of zero for a time step without an input value, which halves the voltage under $\tau$ = 2. The corresponding potential is computed using the time constant and the previous neuronal charge, if an input is present.

\subsection{Spike-Based Character Embedding}
\label{Modeling Textual Embedding Spike-Compatible}
To enable the model to process, a suitable embedding is required. It should not only encode the characters but also position data to retain sequence information. Most often, such information is expressed through a corresponding value array. The desired embedding should generally produce a binary value range in order to conform to the fully spiking design, with one exception, which will be further discussed in Section~\ref{Spike-Compatible Usage of Residuals}. As a first approach, the characters can be encoded as one-hot vectors, meaning they are mapped to a distinct position in an enumerated representation of the alphabet. This fulfills the requirement of a value range between zero and one, but does not yet include positional data and has large space demands, especially when anticipating usage on whole words. One-hot encoding is also unsuitable for displaying similarities and relationships between tokens.

\begin{figure}[htb]%
	\includegraphics[width=\textwidth]{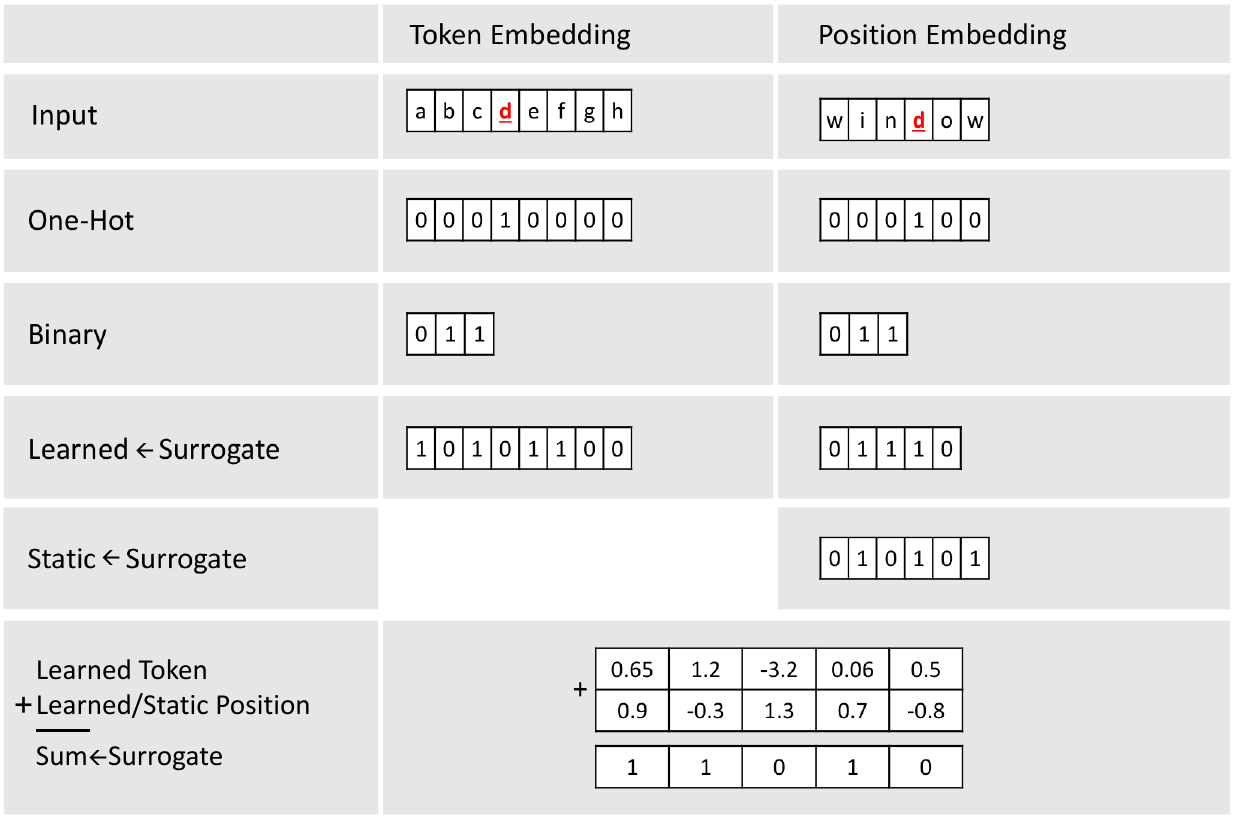}
	\caption{Overview of the different spiking embedding/encoding approaches.}%
	\label{fig:Embedding Approaches}%
\end{figure}

To encode the input position, one-hot encoding can also be used, and the two vectors can be combined to yield a representation with two spikes instead of one. The embeddings produced by this method are large and thus only suitable for structural validation. A similar approach is to encode the position in binary form, producing a significantly smaller but denser vector. When it comes to encoding the surrounding information of a specific signal beyond its identity, however, there are more capable alternatives. Following the original Transformer architecture \citep{vaswani2017attention}, we consider learning token embeddings using a sophisticated objective. In addition, we evaluate the original positional encoding, which alternates between cosine and sine values based on the token position in the input. We also consider a learned positional embedding, which often yields similar results but, according to the authors, is less suitable for very long input sequences. For our last approach, we employ float-based learning/encoding and project to binary values.
We have formulated an overview of these alternative variants to one-hot encoding for a structured evaluation:

\begin{itemize}
	\item \textit{Binary}: Representing an absolute position index in a binary value (spike or no spike), common base two encoding is an option. While this approach may be the most size-effective one ($\lceil$$\log_2$$\rceil$ of the vocabulary size or maximum input length), it will likely be difficult for the model to work with. This is due to the volatile character of the corresponding function, which is hard to approximate, especially as compared to the one-hot or static encoding. 
	\item \textit{Static}: The original Transformer structure uses a form of static encoding through sine and cosine values based on the positional index. Although these functions are not well suited for spike-based computing, this can be circumvented by saving the results in a lookup table prior to model deployment or training. The static creation of the table would be suitable for the scope of the model since it saves computational power throughout application.  In order to obtain a binary input value range for the linear layers at the beginning of the decoder block, we employ an additional LIF unit or directly apply the Heaviside step function with a corresponding surrogate. The information loss from the transformation could be at least partially compensated by extending the value vector while retaining the size advantage over standard one-hot encoding.
	\item  \textit{Learned}: We can employ a common embedding objective to learn a representational position or token vector consisting of floating-point values to express the index or character in the input sequence. Similar to the static approach, this would require the use of an LIF neuron or an activation function block for binary mapping.
	
\end{itemize}
Contrary to natural language vocabulary, positions can generally be represented by an ascending enumeration. However, such a fixed encoding could suggest wrong connections when applied to natural language tokens and thus make it difficult for the model to work with. Hence, we disregard the \emph{static} token-embedding strategy in the following. An overview of the application of different approaches is shown in Figure \ref{fig:Embedding Approaches}.

\begin{figure}[htb]%
	\centering%
	%
	% Including .png
	\includegraphics[width=80mm]{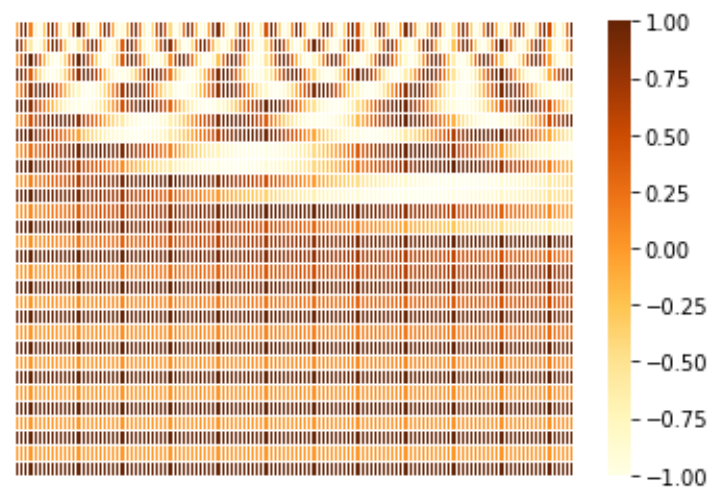}%
	\includegraphics[width=80mm]{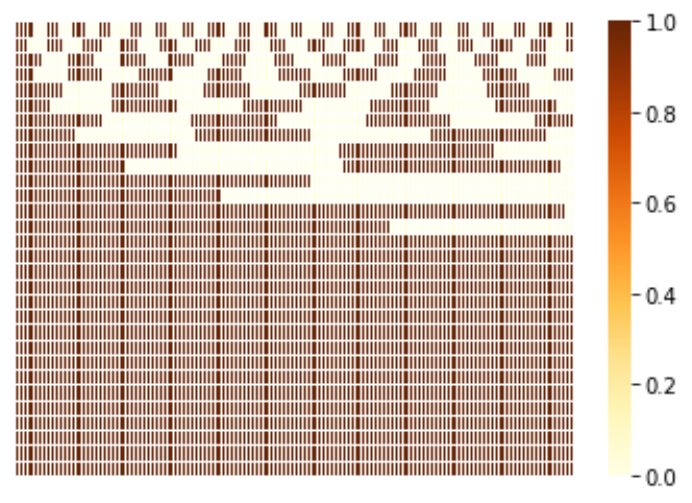}%
	\vspace{0.1cm}
	\includegraphics[width=80mm]{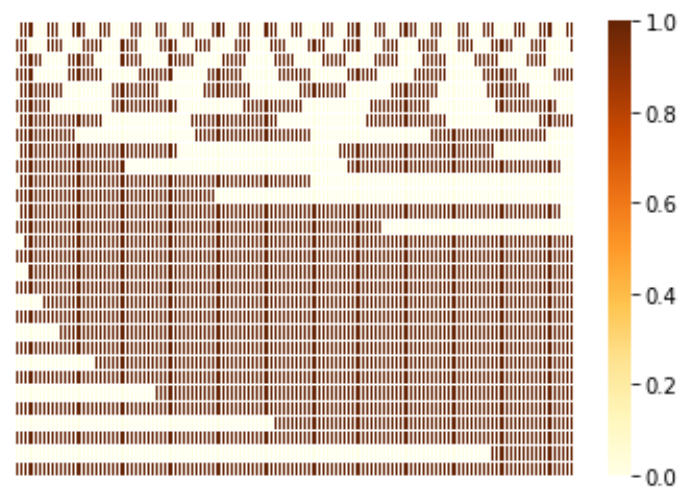}%
	\caption{{Visualization of the static positional encoding, with the position up to 128 on the x-axis and the embedding vector of size 30 on the y-axis. The first plot shows the common float encoding used in ANN transformer structures. The second plot shows the corresponding binary mapping through the Heaviside function. A slight bias towards the value 1 and a potential underutilization of higher-index embedding values can be observed. The third plot shows a static encoding with a subtracted bias of 0.02 before the mapping, producing a better distribution of values and a better utilization of higher embedding indices.}}%
	\label{fig:StaticEncoding}%
\end{figure}

Looking more closely at the resulting mapping from the static approach, it is apparent that the encoding is biased towards positive values, resulting in $\sim$75\% (static encoding of 256 positions using an encoding vector of size 30) ones/spikes due to the Heaviside function. This suggests an inability to make optimal use of the position vector length, since this would result in approximately equal numbers of ones and zeros. However, mitigating this effect by subtracting a small positional bias resulted in insignificant changes in model performance, so we do not consider this correction further. The results of mapping the static position vectors to binary values, both with and without bias, are demonstrated in Figure \ref{fig:StaticEncoding}. 

\subsection{Normalization}
\label{Employing Normalization}
Using normalization layers to rescale activations is a common practice in ANNs to stabilize training and improve generalization. For the purpose of this model, normalization seems necessary, as output values of the matrix dot product in the decoder block range to a maximum of $emb_{dim}*seq_{len}$ if the following LIFs produce a lot of spikes, which is undesirable.
Firstly, we consider Batch Normalization (BN), which is a common technique that has been used in similar models (\cite{zhou2022spikformer}). Alternatively, we also consider Layer Normalization (LN), which is a natural match for natural language applications. In contrast, Batch Normalization may pose issues, as it is a poor choice for processing natural language and has been shown to degrade performance, especially for Transformer-based models (\cite{wang2022understanding}).
As a third option, we evaluate Power Normalization (PN) (\cite{shen2020powernorm}) to address this issue, which aims to provide a Transformer-compatible version of BN.
In addition to performance, it is essential to factor in SNN compatibility. This, and other differences between these three normalization approaches and their respective shortcomings, will be further discussed in \ref{Combining Linear Layers and Normalization for Deployment}. Prior to that, we identify some issues with avoiding multiply-accumulate floating point operations that are inherent to normalization in the next subsection.

\subsection{Spike-Compatible Usage of Residual Connections}
\label{Spike-Compatible Usage of Residuals}
\begin{figure}[htb]%
	\includegraphics[width=\textwidth]{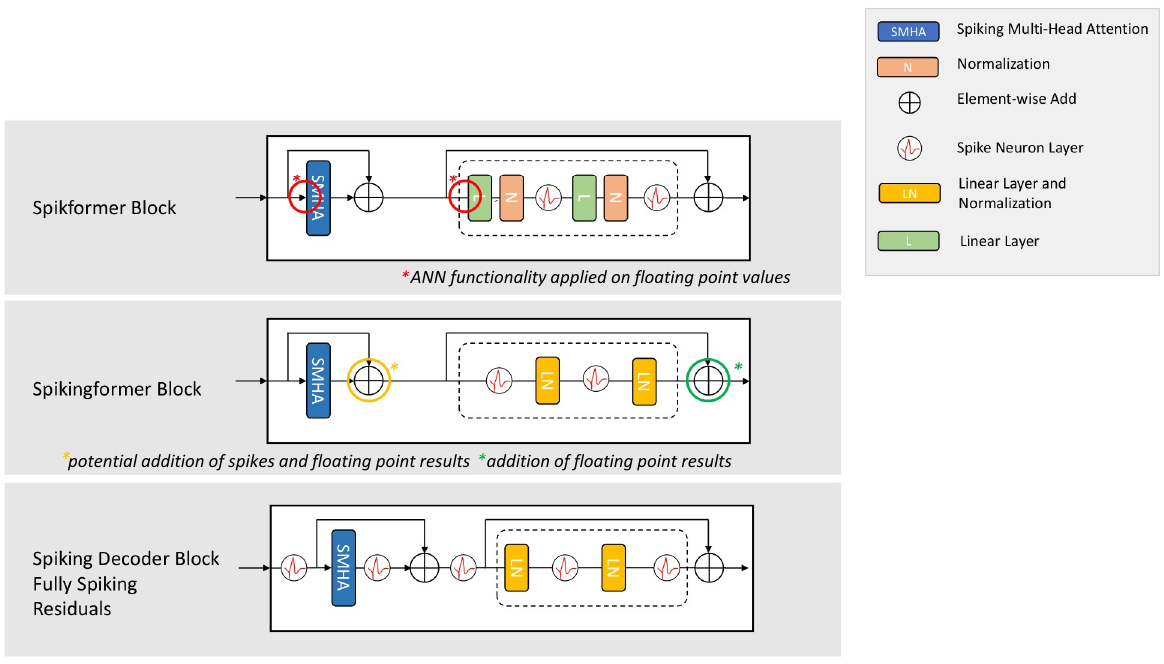}
	\caption{Different residual structures in Spikformer, Spikingformer, and the pure spiking variant which will be tested in this work. In general, this model will, however, rely on structure 2. In addition to the depicted changes, an internal LIF is moved from the end to the beginning of the SMHA from the Spikformer to the Spikingformer block.}%
	\label{fig:Residual Structure}%
\end{figure}

The Transformer architecture includes residual connections around multi-head self-attention and feedforward blocks, which need to be adapted to appropriately handle spikes.
A straightforward implementation proposed in Spikformer~\citep{zhou2022spikformer} is shown in the first row of Figure~\ref{fig:Residual Structure}. It relies primarily on spike-conforming calculations that can produce values outside the spike range due to residual connections around the decoder block components. If chained, applying such decoder blocks will result in floating-point multiplications. This is intuitive, since the residual connection combines two value ranges limited to {0,1}, expanding them to {0,1,2} or even further in some cases. Not only do the resulting float operations increase energy consumption, but they also reduce the overall model performance, according to \cite{zhou2023spikingformer}.

We further review the marked points in the Spikformer model where the issue originates. At the beginning of the MLP, floating-point multiplications are applied in the first layer due to the summation operator before it, which can be avoided by reordering the LIF neurons as shown in the second row for the Spikingformer model\citep{zhou2023spikingformer}. Since the output of the SMHA, together with the other element of the residual connection, is now directly mapped to the binary range again, the results of the SMHA block also no longer have to be binary. While a floating-point result makes the following operation a floating-point addition, this is tolerable for the sake of our model. Therefore, the LIF neuron from the end of the SMHA block is moved to the beginning. This also solves another issue, where the first layers of subsequent blocks were applied to the results of the final residual connection of the preceding block, which are again non-binary. Thus, LIF neurons at the beginning of the MLP and the SMHA ensure that the input data for the following ANN neurons are binary. This removes all floating-point multiplications caused by the residual connections. \cite{zhou2023spikingformer} adapted this new structure to improve energy consumption and overall performance of Spikformer. The resulting structure is depicted in the second row of Figure \ref{fig:Residual Structure}.

The restructuring also enables modifications to the embedding block, which we exploit to evaluate the performance impacts of using float instead of spike input embeddings. This is possible since the new LIF at the start of the SMHA now maps a corresponding embedding vector to spikes, regardless of value range. While we include both approaches, floats incur a mixed-type addition in the residual around the SMHA in the first decoder block. The corresponding point is marked in the second row of Figure \ref{fig:Residual Structure}. This summation is undesirable, and we show that it can be avoided by a different placement of the first input value of the residual connection in the first block.

However, there is still an issue between the linear layer and the subsequent normalization layer in the MLP and SMHA, since both apply float multiplication, and it would be useless to place an LIF neuron between them, since spikes do not need to be normalized. Because both components apply linear transformations, fusing them in deployment would eliminate the need for additional normalization (cited in \cite{zhou2023spikingformer}). Intuitively, the idea is to combine both linear transformations, redefining the weights of the ANN layer to conform to the normalization principle. 

Despite these modifications, the model still relies on floating-point additions for the residual connections. While this is likely an essential part of avoiding vanishing gradients for optimizing the ANN components, exploring pure spike-based communication is interesting. For this purpose, we evaluate an adapted version of the decoder block in which the addition for the residual connections is applied only directly after the LIF neurons. The corresponding structure is presented in the third row of Figure \ref{fig:Residual Structure} and will be further analyzed in subsection \ref{Structural Adjustments: Spike Performance Trade-Off}. It is important to note that this change does not concern MAC operations, so the previous structure generally remains valid for our purposes.

\subsection{Combining Linear Layers and Normalization for Deployment}
\label{Combining Linear Layers and Normalization for Deployment}
Combining an ANN layer with a subsequent normalization layer in the MLP to avoid MAC operations has already been explored for a convolutional layer followed by batch normalization (\cite{zhou2023spikingformer}). For this work, we present an additional fusion of layer normalization and linear layers, as discussed in Section~\ref{Employing Normalization}. The layers apply a linear transformation of the following nature:

\begin{equation} \label{eq:2.1}
	y_{lin} = w_{Lin} * x_i +b_{Lin}
\end{equation}

And the layer normalization may be expressed as

\begin{equation} \label{eq:2.2}
	\begin{split}
		y_i & = LN_{\gamma, \beta}(x_i) = \gamma\frac{x_i-\mu^l}{\sqrt{\frac{1}{H}\sum_{i=1}^{H}(a_i^l-\mu^l)^2}+\epsilon} + \beta \\
		& = \frac{\gamma}{\sqrt{\frac{1}{H}\sum_{i=1}^{H}(a_i^l-\mu^l)^2}+\epsilon}x_i + \beta - \frac{\gamma*\mu^l}{\sqrt{\frac{1}{H}\sum_{i=1}^{H}(a_i^l-\mu^l)^2}+\epsilon}
	\end{split}
\end{equation}

Where $\mu^l$ is the arithmetic layer mean across all units, $a_i^l$ is the output value of one layer unit, $\gamma$ and $\beta$ are the learnable parameters of the applied linear transformation, and $\epsilon$ a small factor to avoid division by zero. Now, since both apply a general linear transformation, they may be fused for deployment in the form of

\begin{equation} \label{eq:2.3}
	\begin{split}
		y_i & = w_{LN} (w_{Lin}*x_i+b_{Lin}) + b_{LN} \\
		& = w_{LN} * w_{Lin} * x_i + w_{LN} * b_{Lin} + b_{LN}
	\end{split}
\end{equation}

The result eliminates the float multiplications by combining weight and bias vectors. The corresponding fusion would consist of a weight parameter and a bias made up by

\begin{equation} \label{eq:2.4}
	W = w_{LN}*w_{Lin}
\end{equation}
\begin{equation} \label{eq:2.4.5}
	B = w_{LN}*b_{Lin} + b_{LN}
\end{equation}

However, this formulation is not sensible for this normalization approach, since Layer Normalization, unlike Batch Normalization, still dynamically computes the arithmetic mean and standard deviation during inference. Computing them by averaging over past samples during training and freezing them prior to inference, as in BN, would likely lead to similar issues to BN in Transformers (\ref{Employing Normalization}). For the concrete formulation of the corresponding fusion of BN and linear layers, we refer to \cite{zhou2023spikingformer}. Lastly, we evaluate the possibility of employing Power Normalization, which provides a similar functionality to BN by applying a fixed linear transformation during inference. Generally, this technique relaxes some BN operations in order to address named issues and employs a running average. It can be formulated as:

\begin{equation} \label{eq:2.5}
	\psi_B^2 = \frac{1}{B} \sum_{i=1}^{B}x_i^2
\end{equation}

\begin{equation} \label{eq:2.6}
	\hat{X} = \frac{X}{\psi}
\end{equation}

\begin{equation} \label{eq:2.7}
	Y = \gamma \cdot \hat{X} + \beta 
\end{equation}

\begin{equation} \label{eq:2.8}
	\psi^2 = \alpha\psi^2 + (1-\alpha) \psi_B^2
\end{equation}

Which is the definition proposed by \cite{shen2020powernorm}. Here, $\psi_B^2$ represents the quadratic mean of batch B, which will be factored into the running average with factor $\alpha$ (constrained between 0 and 1) in eq. \ref{eq:2.8}. In order to enable sensible backpropagation, they suggest a gradient approximation to circumvent backtracking through the prior iterations. Similar to LN and BN, $\gamma$ and $\beta$ represent learnable parameters of the final linear transformation. As obvious from eq. \ref{eq:2.7}, the normalization function may be combined with a linear layer for deployment in a manner similar to the one described previously in this section.

\subsection{Reuniting Synthetic Time Steps}
\label{Reuniting synthetic time steps}
\begin{figure}[htb]%
	\includegraphics[width=\textwidth]{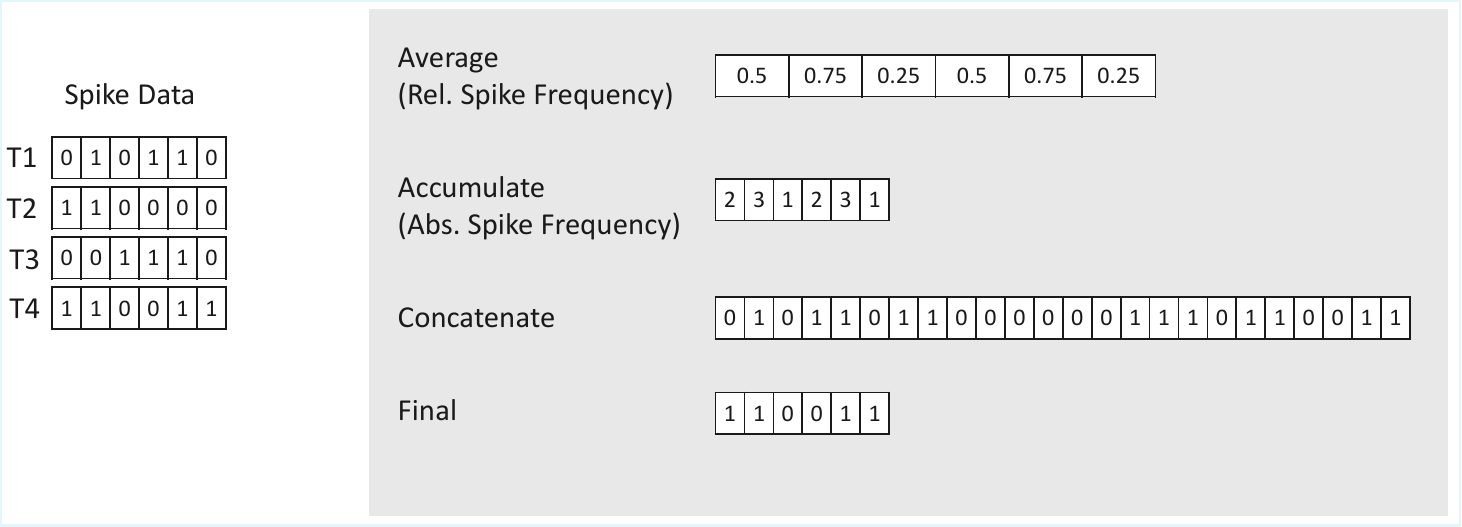}
	\caption{Approaches for the reunion of synthetic time steps. Displayed on the left are the outputs of the respective LIFs for each of the four time steps, and to the right are the corresponding results according to the approach.}%
	\label{fig:UT Approaches}%
\end{figure}

To leverage the event-driven nature of SNNs, additional adjustments to the input structure are needed. The resulting shape is known as neuromorphic or time-varying and includes an explicit time-step dimension. The easiest way to achieve such a shape is through repetition if the corresponding input is static. However, this comes with many arbitrary operations and provides only pseudo-events to the SNN. Regardless, since the model structure is still very new, this strategy is a natural choice, with the option to explore other approaches in the future. Naturally, the LIF neurons consider this dimension when calculating the current potential before integrating the next charge. During inference, there are multiple strategies for deriving data from the time-step dimension. Specifically, we have formulated four approaches:
\begin{itemize}
	\item \textit{Average}: We evaluate the usage of an averaging operation over the time dimension prior to the classification head to aggregate time step information. While this approach is biologically inspired (similar to rate coding), the conversion itself is challenging to design in a spike-compatible manner.
	\item \textit{Accumulate}: A similar approach to averaging, that does not employ the problematic float division, is spike accumulation. Since accumulation comes naturally to SNN models, while the provided information is almost identical to the average, this is a promising approach
	\item \textit{Concatenate}: We present a concatenation operation over the time step results. While very straightforward and without information loss, this strategy requires adjusting the subsequent linear layer size accordingly. This would produce spike-form results and is SNN-compatible, but slightly increases the number of parameters in the head component.
	\item  \textit{Final}: For the final approach, we consider ignoring all but the last results, meaning the model will be trained to produce a spike over $n$ inputs to the LIF, where $n$ is the number of time steps. While this is easy to implement, it inherently produces information loss since spikes before the final step are discarded. Additionally, the neurons' reset behavior inhibits the production of a spike if one was produced in the previous timestep, which may further complicate training.
\end{itemize}
Figure \ref{fig:UT Approaches} shows an overview of the different approaches. The choice of structure can be assumed to correlate strongly with model performance. Therefore, a non-spiking variant that performs linear-layer computations on floating-point values is also considered. For all of these strategies, we further discuss the choice of the location at which reunion is performed and required adjustments to the classification head. \newline

\begin{figure}[htb]%
	\includegraphics[width=\textwidth]{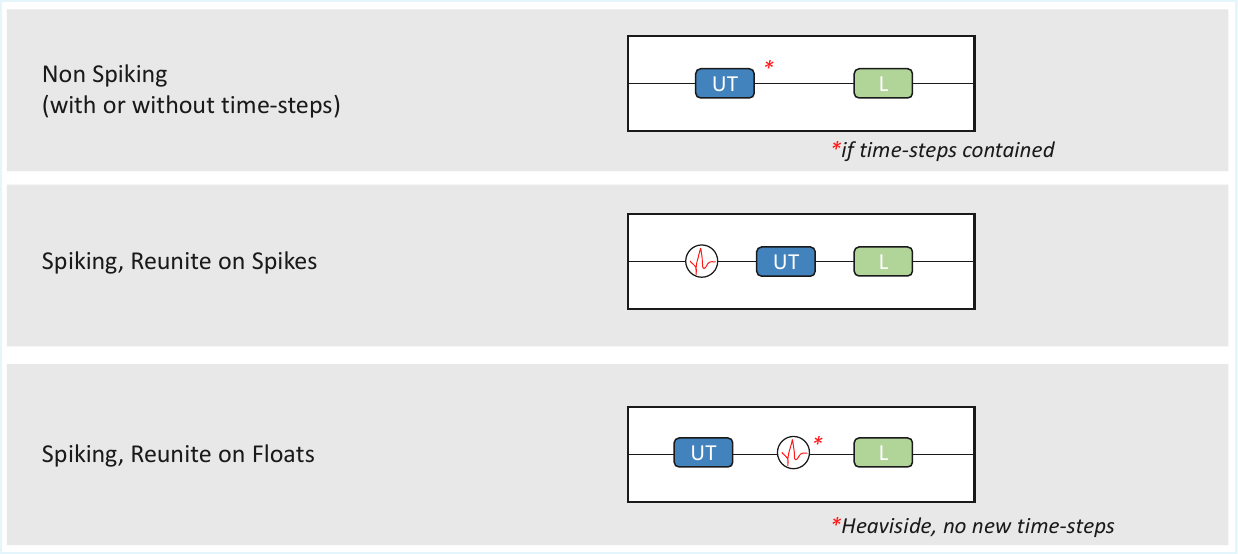}
	\caption{The head structure is dependent on the spiking property of the head, as well as the choice of the unify time steps (UT) object placement.}%
	\label{fig:Head Structure}%
\end{figure}

Furthermore, for the design of the fully spiking variant, which contains an additional LIF, there are multiple options for the placement of the time-step unification block. Both before and after this LIF neuron are valid choices, which are displayed in rows two and three of Figure \ref{fig:Head Structure}. Placing a LIF after the UT block may seem unintuitive, but just constitutes the application of a Heaviside function with a suitable surrogate for backpropagation. This property can be utilized to map the unified data sequence to Spike-form again before applying the linear layer. In general, if not specified otherwise, we use the structure featuring an LIF before the UT block. 

\pagebreak
\section{Results}
\label{Results}
In the following subsections, we first describe the non-spiking baseline decoder architecture before introducing the architecture of our SpikeDecoder model based on the components identified in the last section.
\subsection{Non-Spiking Decoder Model}
\begin{figure}[htb]%
	\centering
	\includegraphics[width=\textwidth]{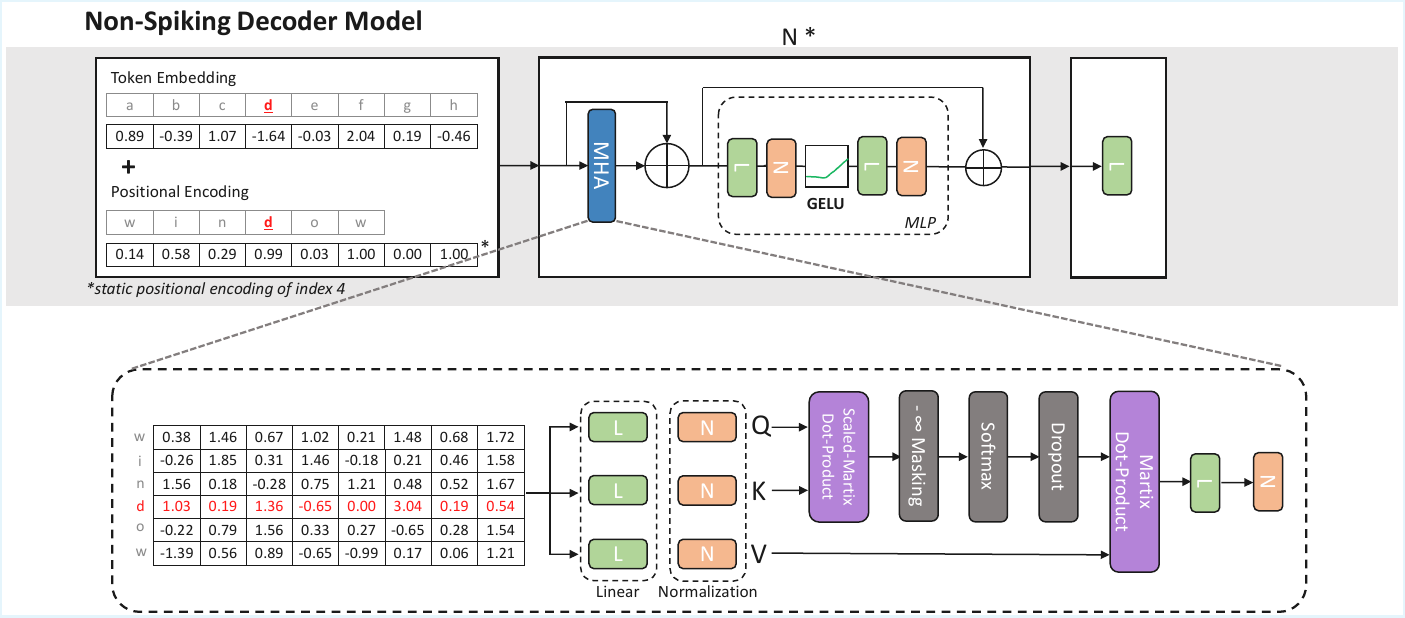}
	\caption{Full non-spiking model architecture, consisting of N decoder blocks, an encoding object and a classification head. Special attention should be paid to the different embedding processes of the input data and the different steps following the QKV operations in the Multi-Head-Attention.}%
	\label{fig:NonSpiking model architecture}%
\end{figure}

Before describing the architecture of the proposed spiking model, we introduce an ANN-based decoder structure. This will serve as a baseline and may be consulted to compare performance later on.
We adopt the decoder-only model architecture proposed by \cite{Karpathy}. In addition to the original attention mechanism, this model uses learned floating-point token embeddings. Also, the aforementioned static position encoding can be directly added (position-wise) to the token embeddings. Besides that, we employ techniques from previous sections that are ill-suited to SNN-based architectures, such as applying softmax in the multi-head attention blocks and including an activation layer (GELU) in the MLP.
The complete architecture of the non-spiking model can be found in Figure \ref{fig:NonSpiking model architecture}. It is also noteworthy that the linear layers with the following normalization are not combined in this scenario. Having revisited the basic model outline for a decoder-only model, we now proceed to the architecture of the fully spiking model, SpikeDecoder.

\subsection{SpikeDecoder Architecture}
\begin{figure}
  \centering

  \begin{subfigure}{\textwidth}
    \centering
    \includegraphics[width=0.9\linewidth]{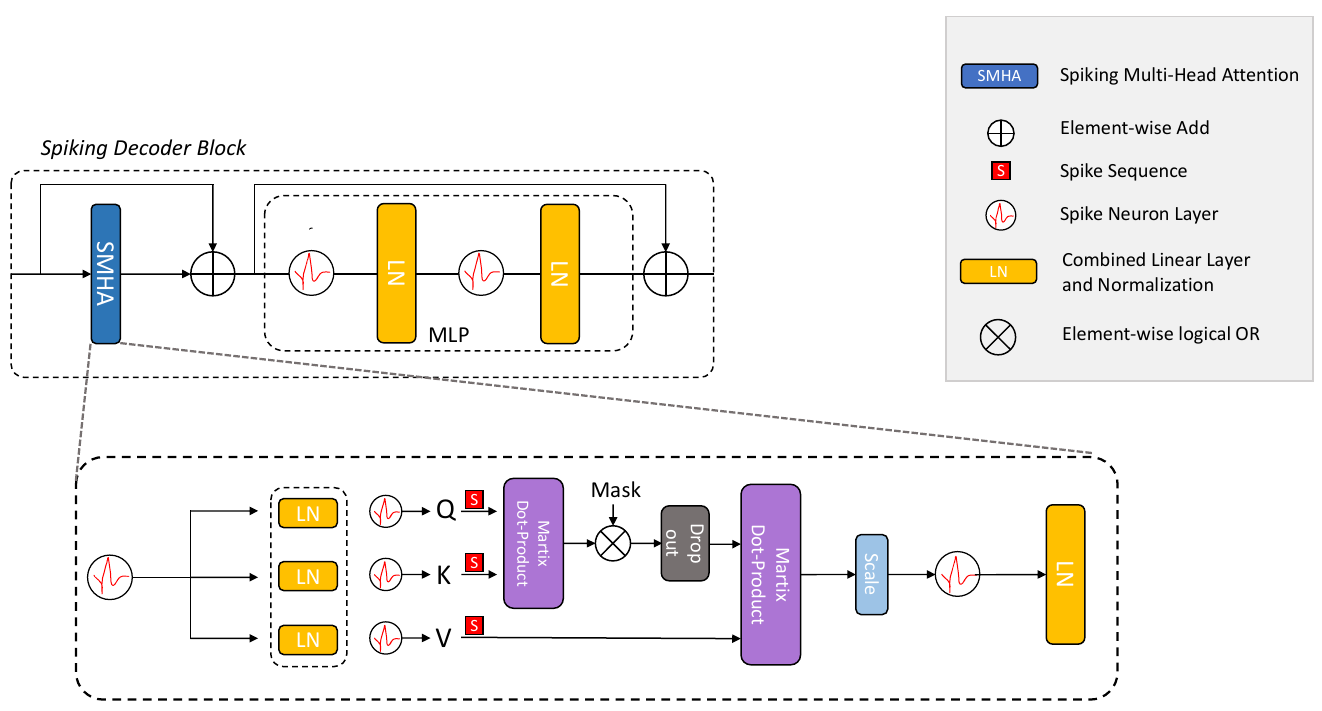}
    \caption{Model architecture of the spiking decoder block. Notable differences to the encoder block of \cite{zhou2022spikformer} are the usage of dropout, the look-ahead mask, and the different choice of normalization.}
    \label{fig:decoder block architecture}
  \end{subfigure}

  \vspace{1em}

  \begin{subfigure}{\textwidth}
    \centering
    \includegraphics[width=0.9\linewidth]{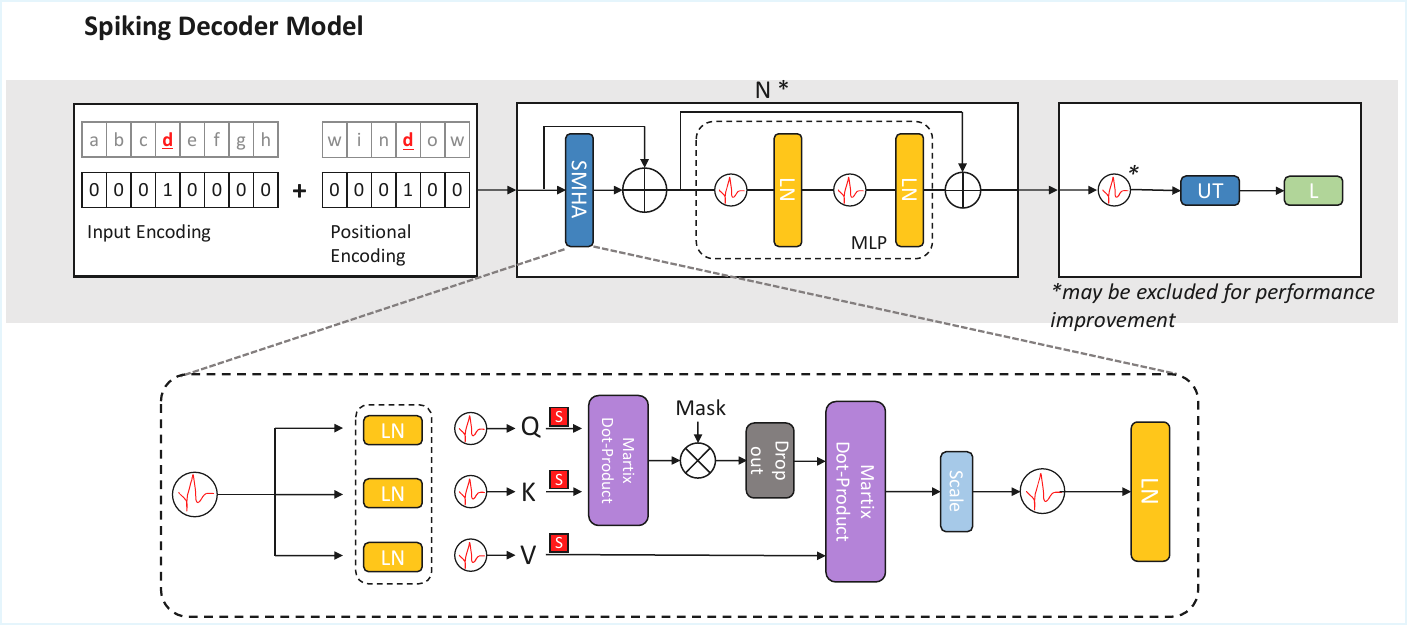}
    \caption{Full model architecture, consisting of N decoder blocks, an encoding object, and a classification head.}
    \label{fig:Full model architecture}
  \end{subfigure}

  \caption{Architectures used in this work.}
  \label{fig: model_subfigures}
\end{figure}

In comparison with the non-spiking version, we need multiple alterations to support spikes. Firstly, as subsection \ref{Combining Linear Layers and Normalization for Deployment} explained, we merge normalization with the preceding linear layers. Similarly, we adjust the look-ahead mask and remove the softmax operation. Lastly, we employ the SpikingJelly Multistep LIFs. An overview of the isolated spiking decoder block can be seen in Figure \ref{fig:decoder block architecture}. Furthermore, we employ the spike-compatible text embedding, as explained in Section \ref{Modeling Textual Embedding Spike-Compatible}. Concerning the residual connections, it is worth noting that the model employs directly following spiking neurons to project results to a binary value range again, as was proposed in \cite{zhou2023spikingformer}. Finally, we insert a LIF, followed by a unification time step (UT) object, into the classification head before the final linear layer. As discussed in \ref{Reuniting synthetic time steps}, the spiking neuron in the classification head may be omitted for performance improvement. Similarly, the position and strategy of the UT object may be shifted. The full model, including embedding and classification head, is displayed in Figure \ref{fig:Full model architecture}.

Notably, the model does not output probability values since softmax or other variants are not spike-compatible. This prevents the structure from employing probability-based search algorithms like Beam Search over the generated tokens, which is generally associated with improved result quality. Similarly, we cannot regulate the sampling temperature. In the future, exploring how such features can be suitably realized in SNNs may be interesting.

\subsection{Partially Spiking decoder Models} \label{Partially Spiking decoder Model}
\begin{figure}%
	\centering
	\includegraphics[width=\textwidth]{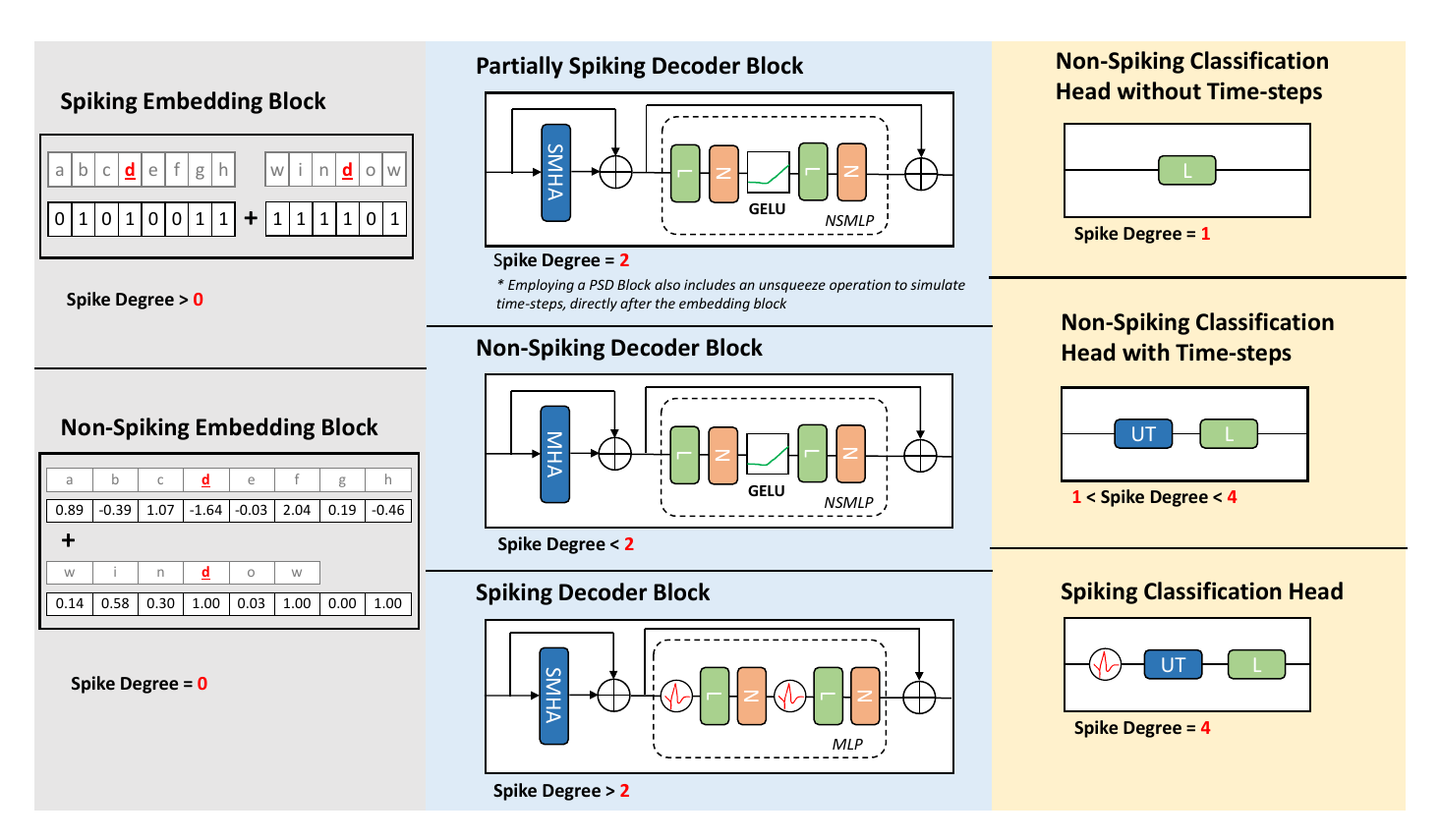}
	\caption{Overview of the different PSD variants separated by the spiking degree. The model is composed of one element per colored sector (blocks are still chained). In addition to the non-spiking and spiking blocks, this structure incorporates partially spiking decoder blocks, which contain spiking multi-head attention and a non-spiking MLP. For the transition, we disregard the combination of linear layer and normalization.}%
	\label{fig:PSD structure}%
\end{figure}

Several differences exist between the spiking and non-spiking decoder models. To investigate further whether there is a decrease in performance and, if so, where it originates during the transition to spikes, it is a sensible next step to implement a hybrid model that includes both non-spiking and spiking components. This partially spiking decoder model can be structurally separated into four major parts: the embedding, the SMHA, the MLP, and the classification head. While the embedding block functionality may not appear as a noticeable difference since the position of the first LIF in the SMHA also enables float embedding for the spiking decoder model, for simplicity's sake, this may be treated as a delayed variant of binary embedding. Secondly, it is necessary to establish a partially spiking decoder block to separate multi-head attention and MLP from an implementation perspective. A visualization of the structures that correspond to the different spike degrees is shown in Figure \ref{fig:PSD structure}. The spike degrees can be categorized in the following manner:
\begin{itemize}
	\item \textit{Spike Degree 0}: Formally equal to a non-spiking decoder model.
	\item \textit{Spike Degree 1}: Common non-spiking embedding block is replaced by a spiking embedding block. The strategies can still be customized, but in general, learned token and static positional encodings are used to provide a natural transition. Unlike the usual spiking embedding, this degree does not provide a time-step dimension.
	\item  \textit{Spike Degree 2}: Float-based multi-head attention is exchanged with the SMHA component. To combine it with the non-spiking MLP, this model includes a partially spiking decoder block. Since this is the first model to feature SNN nodes, the embedding block generates synthetic time steps. Therefore, it is also necessary to include a UT object in the classification head to join the time step information before the final prediction.
	\item  \textit{Spike Degree 3}: The partially spiking decoder block is replaced with a fully spiking one. This change replaces the non-spiking MLP with the one used in the fully spiking decoder model. 
	\item  \textit{Spike Degree 4}: The final degree includes an LIF node in the classification head. The placement of the unification process also remains unchanged. This model is equal to the fully spiking decoder model (SpikeDecoder).
\end{itemize}
With these degrees, we can effectively analyze the performance fluctuations in the transition process and other potential side effects. 

\pagebreak
\section{Evaluation}
\begin{figure}[p]%
	\centering
	\includegraphics[width=\textwidth]{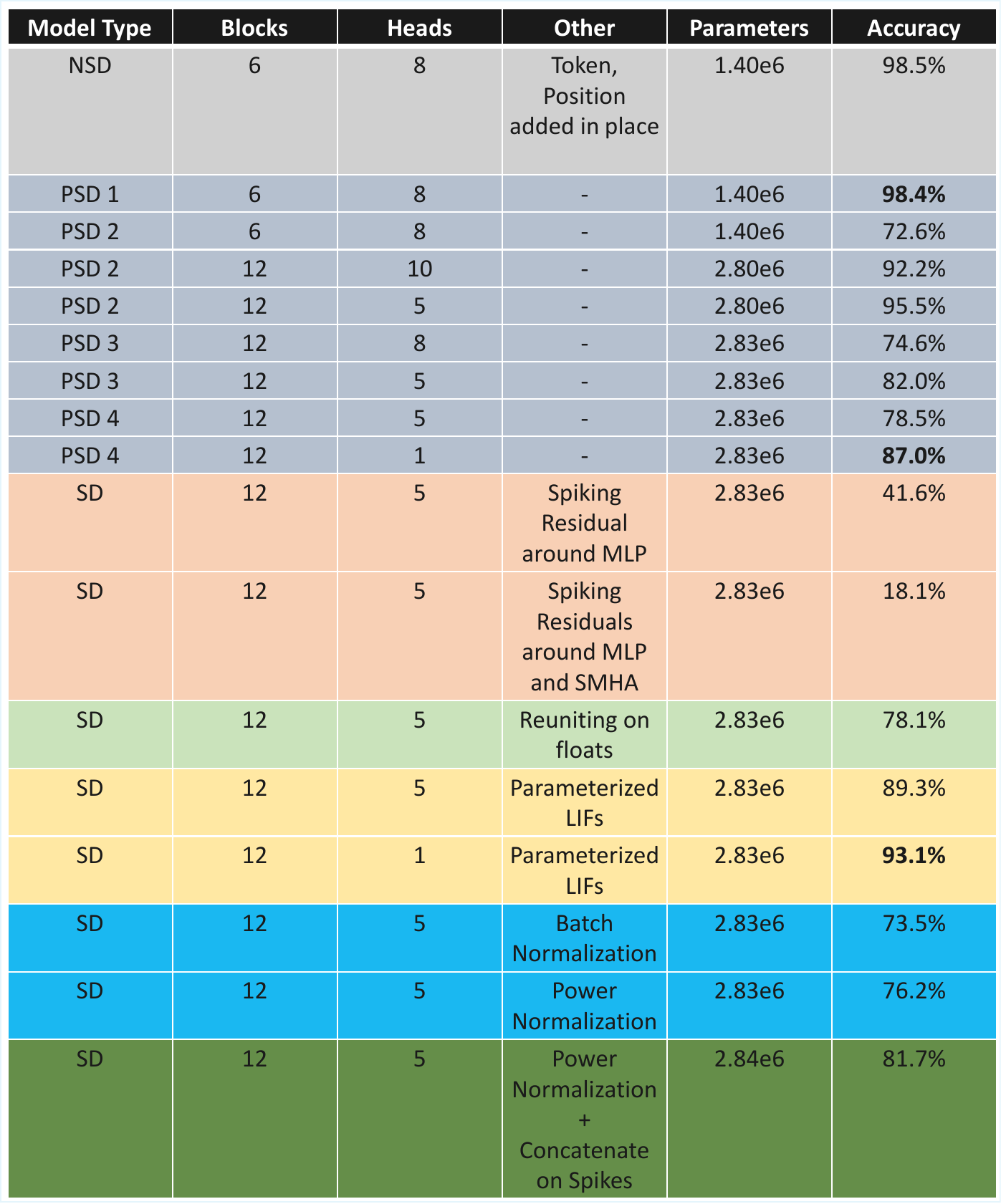}
	\caption{Comparison of the train set accuracy of different model variants. ND refers to non-spiking decoder model, PSD to partially spiking decoder model with the respective degree and SD refers to the fully spiking decoder model (equal to spike degree 4).}%
	\label{fig:Performance Comparison}%
\end{figure}

In this section, training results will be discussed for both spiking and non-spiking models, the specifications during the transition to the fully spiking variant, and the effects of further structural changes. We also estimate the theoretical energy consumption. For a swift comparison of different model applications, we provide a synoptic results table in Figure \ref{fig:Performance Comparison}.
For general performance measurement, we use prediction accuracy relative to the source text training set. Since the model is supposed to learn general language generation, this is not an optimal choice, but it is intuitive for the current scope.
It is also important to note that readable text outputs are considerably more difficult to generate for this model since there is no predefined word vocabulary to refer to. In addition, the model does not produce probability outputs, meaning it cannot fall back on corresponding search algorithms. In turn, a single mistake in earlier predictions can result in greater difficulties in later sections.

\subsection{Non-Spiking decoder Performance} \label{Non-Spiking decoder: Application on Natural Language}
For testing the model on natural language, we use ``War and Peace'' by Leo Tolstoy. Due to the size, the train set is shortened to make training comparisons possible in a feasible time. The data is reduced to a 100,000-character-long excerpt, combined with a 90/10 train/test split and a random sampler. The input length is set to 256, enabling the model to generate about 1.5 to 2 sentences with a reasonable input prompt.

The model is trained using an embedding dimension of 80 (combined token and positional encoding), eight attention heads, and six chained decoder blocks. This results in a parameter number of about 1.4 million and is chosen to enable a suitable comparison through the transition process. Using this configuration, the model achieves 98.5\% accuracy after a short training period.
Generally, the model performs about as well as should be expected. It generates suitable text excerpts from input prompts that are similar to the source text's style and content. However, due to the comparatively small training set, even the ANN model struggles to operate on texts that are far from the source text. This may imply increased performance in bigger use cases later on.

Having established a stable performance baseline, we begin preparing for the transition to spikes. In turn, the different embedding options proposed in Section \ref{Modeling Textual Embedding Spike-Compatible} will be formally compared next. While the non-spiking decoder uses a position-wise addition of learned token embeddings with static positional encoding, there is little to no research on the most effective form of embeddings for spiking Transformers in natural language processing. This is a natural next step to prepare for the training of spike degree one.

\subsection{Embedding Performance Comparison}
We generally consider both spike and floating-point embeddings, due to the reordering discussed in Section \ref{Spike-Compatible Usage of Residuals}. Since the choice of embedding strategies inherently influences the model's size, it is rather difficult to produce an objective comparison. Although extending the length of binary or other static embeddings by padding would be possible, this would be counterintuitive, since their biggest advantage is their reduced size. In a similar manner, the one-hot encoding approach produces a large model, which, in turn, improves performance.

\begin{figure}[htb]%
	\centering
	\hspace{-0.5cm}
	\includegraphics[width=81mm]{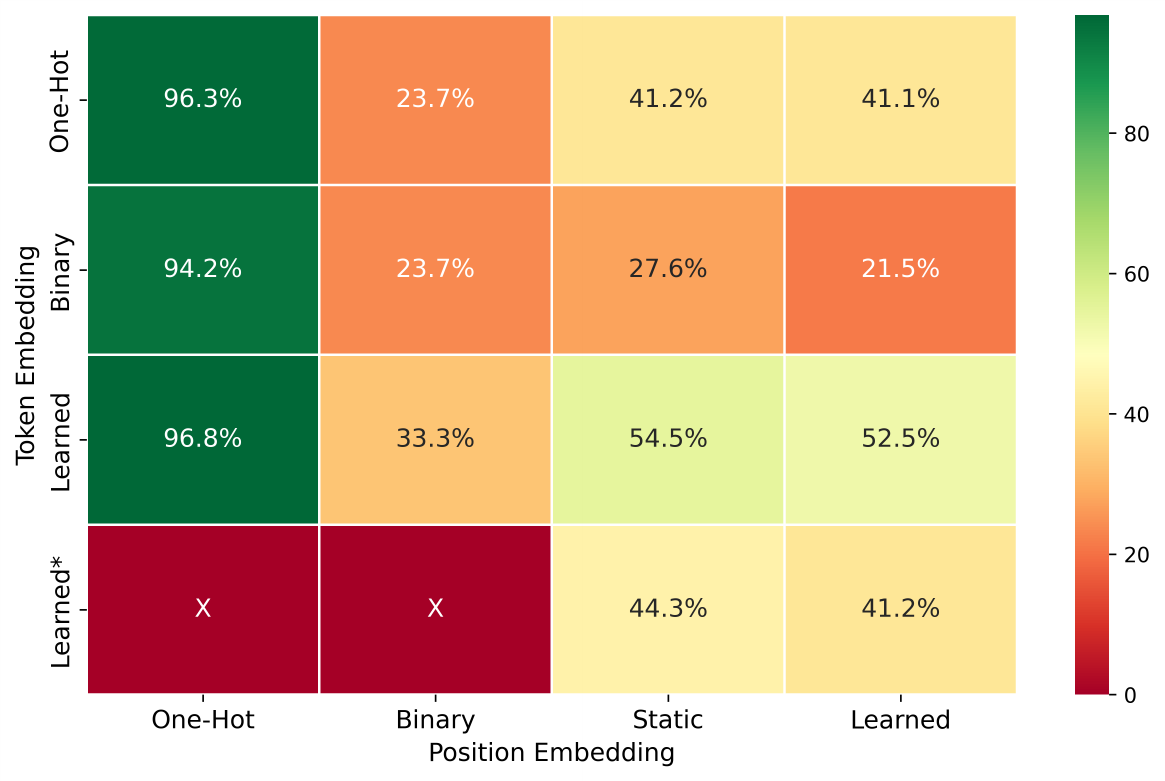}%
	\hspace{0.5cm}
	\includegraphics[width=81mm]{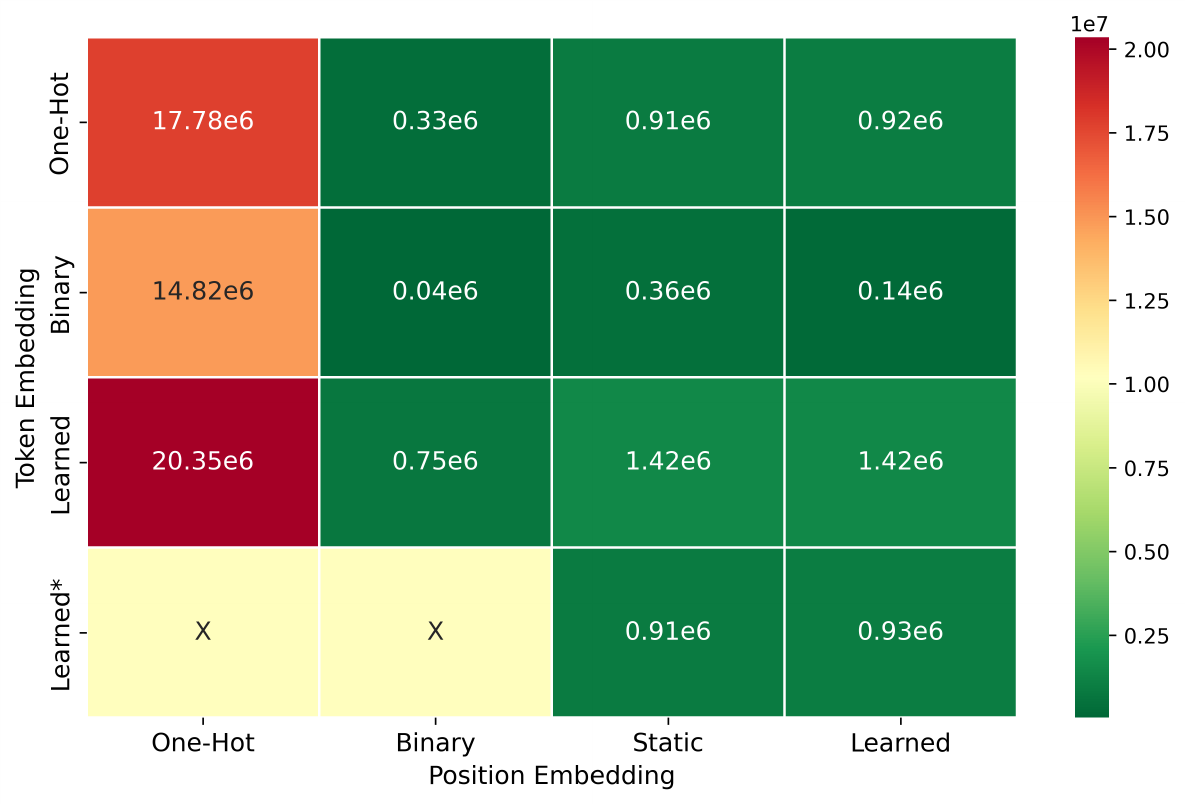}%
	\caption{The left table displays the accuracy of next character prediction given the respective combination. The right table shows the number of parameters, and is strongly dominated by the combinations with linear positional encoding. The token embedding is shown on the y-axis, and the position is on the x-axis. In the last row (marked with $\text{Learned}^*$), the position and token embeddings are added, which is why one-hot and binary position embeddings are excluded.}%
	\label{fig:Embedding_Comparisons}%
\end{figure}

The accuracy overview in Figure \ref{fig:Embedding_Comparisons} clearly shows significant disparities across different combinations, partially due to size differences. For the following sections, however, a couple of approaches can be eliminated. The One-Hot approach is not suitable for modeling large input sizes, as it leads to an explosion in model parameters. On the contrary, binary embedding is very well suited for this task but almost halves the average overall performance, potentially also due to the reasoning mentioned in Section \ref{Modeling Textual Embedding Spike-Compatible}. Learned and static approaches for the positional embedding-only feature show only minor performance differences between them. While static embeddings are computationally less demanding, learned embeddings may adjust better to the subsequent mapping to binary values, which could be useful since static encoding was developed for floating-point representations. Also, a decision must be made on whether to concatenate or to add the token and position embeddings. While concatenating the two increases the input size, it may be difficult for the input to preserve positional information when computing the sum.

In addition to the combinations in the comparison table, we have considered delaying the projection to the binary range until the first LIF unit in the SMHA block. Surprisingly, this approach did not improve the performance of the top-performing combination with a reasonable size and, on average, even slightly decreased it by about 1.
Overall, the concatenation of learned token embeddings and static positional encoding showed the greatest promise, with moderate and dynamic size requirements and very stable performance. Static position encoding also provides dynamic size allocation and already has a solid theoretical foundation.

\subsection{Transitioning to Spike-Form: Partially Spiking Decoder Models}
\label{Transition to Spike-form, Partially Spiking decoder Models}
To identify the pain points for the new model, the structure will be gradually transitioned to use binary values, employ LIF neurons in the Multi-Head-Attention, and finally eliminate all MAC operations within the model.
Correspondingly, we investigate the partially spiking model architectures presented in Section \ref{Partially Spiking decoder Model} and their respective performance. Although float embeddings have not shown any significant performance increase in the previous sections, we employ them to avoid the mixed-type addition mentioned in Section \ref{Spike-Compatible Usage of Residuals}. Since spike degree 0 is equivalent to the fully non-spiking decoder model, we do cover it again. Throughout the transition, we will directly address and incorporate factors to compensate for performance drops. This is done to arrive at a final result as close as possible to the ANN model's capacity.

In transitioning from a non-spiking to a spiking model, we observe nuanced changes in performance across the different spike degrees.
For Spike Degree 1, using a spiking embedding block yields comparable loss and accuracy, with test prompt outputs of similar quality. Notably, a reduction of the token embedding from 80 to 50, as well as the shift to binary values, does not significantly impact performance, as demonstrated in Appendix Figure \ref{fig:Token_Embeddings}. The new binary embedding retains a clear, unique mapping from characters to vectors that features similarities in shorter sequences for more closely related characters, such as 'u' and 'v'.

Moving to Spike Degree 2, incorporating spiking multi-head attention results in the most significant performance decline of 25\%. Strategies such as increasing the amount of decoder blocks and reducing the number of attention heads partially mitigate this decrease, achieving an accuracy of 92.2\%.

In Spike Degree 3, replacing the MLP leads to a 13\% accuracy decrease, but the model retains acceptable performance, especially in comparison to non-spiking counterparts. Finally, introducing a spiking classification head in Spike Degree 4 causes a 4\% accuracy drop, which can be compensated by reducing attention heads to one, resulting in a final accuracy of 87.0\%. Adjustments in model accuracy throughout the transition are illustrated in Appendix Figure \ref{fig:Transition Graph}.

Overall, while performance decreases are observed in transitioning to spiking components, strategic adjustments mitigate these effects to create a functional SpikeDecoder model instance, which performs around 11.4\% worse than the used ANN counterpart.

\subsection{Design Exploration and Performance Trade-Offs} \label{Structural Adjustments: Spike Performance Trade-Off}
In addition to the exploration of the transition from an ANN-based on an SNN-based model, several components of the SpikeDecoder architecture may further influence the model performance. For example, as shortly addressed in Section \ref{Reuniting synthetic time steps}, one may consider the different positions of the UT object and the corresponding unification strategies.

Also, when investigating the fully spiking model, it is sensible to examine the importance of the floating-point addition residual connections. Presumably, they constitute a vital part of the optimization process, since the LIF neurons might otherwise cause vanishing gradients during backpropagation. However, eliminating the corresponding floating-point additions would reduce the model's overall energy consumption and make it a native SNN. The structural approach to realize this change consists of adding one LIF directly behind the SMHA and one at the end of the MLP. Additionally, we adjust the position of the residual connection by pulling the respective internal LIFs in front of the connections around SMHA and MLP. The resulting design is shown in the third row of Figure \ref{fig:Residual Structure}.

Moreover, we separated the normalization type from the transition process for simplicity's sake. Regardless of this choice, it is necessary to replace the default layer normalization with a pure spiking model, as discussed in \ref{Combining Linear Layers and Normalization for Deployment}. Comparing the potential candidates of Batch and Power Normalization is another interesting point to consider.

Lastly, the current LIF implementation is static, meaning properties such as spiking threshold or reset value cannot be adjusted during backpropagation. Registering either of them as trainable parameters may increase the model's flexibility. For all the following experiments, we use final model of the transition process, but employ five attention heads instead of one.

\subsubsection{Residual Connection}
First, we investigate a pure spiking residual connection structure in the decoder block, meaning we solely add binary values in the surrounding add connections. We have already discussed the resulting structure (Figure \ref{fig:Residual Structure}, bottom row) and will compare it with the one with floating-point-based residuals.
We separate the two residuals of the structure according to the component they surround for testing purposes. Introducing the residual connection only around the MLP already reduces performance to 41.6\%. This is only about half of the original accuracy and again shows the vital role that the residuals play in model optimization. Adding the new structure around the SMHA model further worsens the final accuracy to 18.1\%. The model evidently depends on residual connections to compute valid gradients, which are needed for functional backpropagation. In turn, this also indicates that the spiking residual connection structure proposed here is not a suitable approach for designing the model's communication in a more spike-driven manner.

\subsubsection{Timestep Unification}
Multiple approaches exist to join the time-step information prior to classification. First, we compare the different strategies employed in the unification object (see Figure \ref{fig:UT Approaches}). After reduced training (25 instead of 50 epochs), the strategies result in the following accuracies: average: 71.5\%, accumulate: 72.2\%, concatenate: 70.0\%, and final: 63.2\%. Intuitively, the \textit{final} strategy restricts the available information to the model during inference, resulting in lower accuracy. The similarity in the results of the other strategies can be explained by the fact that their results convey essentially the same information.

Next, we investigate the placement of the UT object inside the classification head. Commonly, the model reunites the temporal dimension of spike sequences. This means that the output of the final decoder block is first projected to spike range through a LIF node and then combined via the unification object. Alternatively, we consider applying the UT block directly to the output of the decoder block. For strategies besides \textit{average}, this is spike-compatible. However, since the data is now floats, another LIF is needed after this unification. To avoid creating new time steps, this node will serve only as an adjusted Heaviside function without integrating over time. The non-spiking structure, including a UT object, was already tested indirectly in spike degree three. The spiking structure that reunites on floats can be reviewed in the third row of Figure \ref{fig:Head Structure}.  
Using this head structure, combined with the averaging strategy, the common SpikeDecoder configuration achieves 78.1\% accuracy. As a reminder, the corresponding performance with the spike-based unification structure using the \textit{average} strategy achieved 78.5\%. Since reuniting on spikes without either the \textit{final} or \textit{concatenate} strategy produces floating-point values for the final linear layer to work on, it is unsurprising that this version performs a little better, but the difference is generally negligible. Overall, we conclude that the unification strategy has a limited influence on total performance, except for the \textit{final} strategy. 

Besides performance, the strategy's suitability for SNN-based implementation must be considered. \textit{accumulate} can be modeled using spiking components but produces non-spiking values. \textit{concatenate} and \textit{final} can be modeled in this manner as well, but produce spike-values, which makes them suited for the placement reuniting on spikes. Since the \textit{final} strategy has been observed to reduce performance, \textit{concatenate} appears to be the most suitable candidate for a pure spiking architecture that reunites on spikes. For the alternative positioning, the unification doesn't need to produce spike values, meaning \textit{accumulate} and \textit{concatenate} are both valid approaches. For the pure spiking variants in this work, we use the strategy \textit{concatenate} on spikes since it enables the classification head to operate on absolute spike frequency.%, and this approach is biologically supported.

\subsubsection{Parameterizing LIFs with a Learnable Threshold}
\begin{figure}[htb]%
	\centering
	\includegraphics[width=\textwidth]{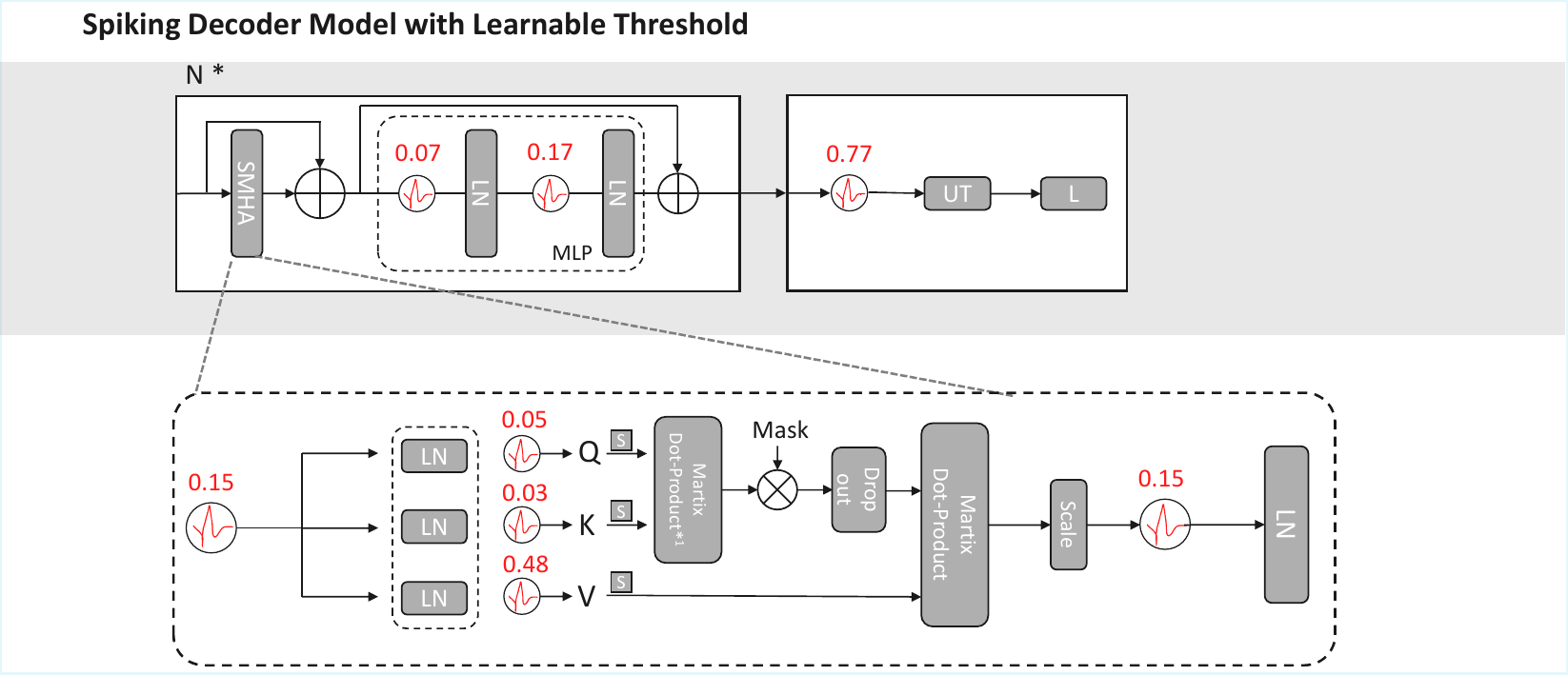}
	\caption{Average learned threshold values for respective LIFs throughout the decoder model.}%
	\label{fig:Threshold Values}%
\end{figure}

Our chosen LIF neuron model has several parameters that are set to static values by default. In the following, we investigate learning the spike threshold parameter.
An experiment with the final transition model results in a performance increase to 89.3\%. Investigating the resulting model further, there are a couple of interesting properties about the new thresholds. Figure \ref{fig:Threshold Values} depicts the arithmetic mean values over the twelve blocks. It is apparent that all threshold values have been decreased from their default value of one. The first MLP LIF and the respective key and query LIFs have particularly low spiking thresholds. When looking further into the different block values, this instance also reveals processing flaws regarding the biological motivation of the LIF neuron. This is due to some key and query LIFs in middle blocks, which now feature a negative threshold value. Most likely, this is caused by the inputs, which are normalized but not restricted to positive values. While this is not problematic for the application of the model itself, biological neurons are not meant to receive negative currents. Application-wise, this could be addressed by restricting the value range of the linear transformation parameters, but this is beyond the scope of this work. Generally, the dynamic nature of the new LIFs appears to enable the model to further tailor the functionality of certain parts. Correspondingly, \textit{query} and \textit{key} generate spikes very easily, while the \textit{value} is rather conservative. Similar behavior may be observed in the MLP, where thresholds are much more diverse across blocks.

\subsubsection{Normalization}
As discussed in Section \ref{Combining Linear Layers and Normalization for Deployment}, Layer Normalization is not suited for the fully spiking decoder model. Despite this, we used it by default for the prior experiments since it speeds up training and can easily be incorporated implementation-wise. Now, we investigate possible effects on the performance when replacing layer normalization with Batch (BN) or Power Normalization (PN). Generally, one can observe that both BN and PN take about twice as long as Layer Normalization during training. This may be partially due to the computation of running averages and the additional reshape operations needed to incorporate BN.

Expectedly, the Batch Normalization has a negative effect on the overall model performance. The accuracy decreases to a value of 73.5\%, which corresponds to a performance loss of about 5\%. This performance drop is undesirable, making Batch Normalization a subpar choice for SpikeDecoder.

Power Normalization, however, appears to have a less significant effect on accuracy, which decreases to 76.2\%. Since this normalization type prevents the fusion of linear layers with their corresponding normalization after training, it is an intuitive choice for the pure spiking variant. Compared to Layer Normalization, the biggest disadvantages are the increased training time and the higher memory requirements, which have to be tolerated in this case.

After investigating these additional structural adaptations, which enable the design of a fully spiking model, namely Power or Batch Normalization, and one of the spike-compatible time-step unification strategies, we want to validate the performance of SpikeDecoder. After training with Power Normalization and reuniting on spikes with the \textit{concatenate} strategy, the model reaches a stable accuracy level of 81.7\%. This successfully demonstrates SpikeDecoder's ability to operate at the purely spiking level without employing any floating-point MAC operations.

\subsection{Theoretical analysis of the Model's Energy Consumption}
\label{Theoretical Energy Consumption}
After analyzing the performance merits of the model thoroughly, we present the estimated energy savings. In a similar manner to \cite{zhou2023spikingformer}, we estimate by distinguishing between FLOPs (multiply-and-accumulate of floating point values) and SOPS (theoretical synaptic operations). Generally, we estimate the number of operations of a matrix dot-product between A (a,b) and B (b,c) using:
\begin{equation} \label{eq:4.0}
	FLOPs_{A,B} = (2b-1)ac
\end{equation}
Which we adapt for a sparse, spike-form matrix through
\begin{equation} \label{eq:4.1}
	SOPs_{A,B} =  T * 2*nnz_A*c
\end{equation}

We use a slightly different estimation than the authors of Spikformer, who directly multiply using the respective neuron firing rate. Here, $nnz_A$ refers to the non-zero values in matrix A, and T to the time-steps created through static repetitions. For estimation purposes, we have computed average rates for the respective matrices in the multi-head attention and the subsequent Multi-Layer Perceptron. Depending on the number of time steps, the values generally range from 10\% to 36\%. Assuming MAC and AC (SOPS) to be implemented on 45nm equipment according to \cite{6757323}, we calculate results with energy consumptions $E_{MAC} = 4.6pJ$ and $E_{AC} = 0.9pJ$. The theoretical consumption of a SpikeDecoder block can thus be formulated as \newline

\begin{equation} \label{eq:4.2}
	E^{DecoderBlock} = E_{AC} * (SOPs_{SMHA} + SOPs_{MLP})
\end{equation}

\begin{equation} \label{eq:4.3}
	SOPs_{SMHA} = T*(6*nnz_{in}*E + 4*S*E+2*nnz_{Q}*S+2*nnz_{QK}*E+2*nnz_{Atn}*E)
\end{equation}

\begin{equation} \label{eq:4.4}
	SOPs_{MLP} = T*(2*nnz_{i}*dim_{hid}*E+S*E*dim_{hid} + 2*(nnz_{o})*E+S*E)
\end{equation}

Where S is the sequence, E is the embedding dimension, $nnz$ the corresponding amount of nonzero values and $dim_{hid}$ the hidden dimension in the MLP. For simplicity's sake, we disregard the separation into different attention heads.  Both the application of a linear layer (after parameters have been fused with normalization) and the two main parts of the spiking self-attention can be formulated as a matrix multiplication (and a following element-wise addition for the layer). 
It is worth noting that this is the fully spiking variant. In addition, the assumptions about LinearNorm only hold for normalization of type BatchNorm or PowerNorm. The corresponding reasoning has been addressed in \ref{Combining Linear Layers and Normalization for Deployment}.

Now, to illustrate the distance between the non-spiking decoder and the SpikeDecoder block, we compare the theoretical energy consumptions.
For this purpose, we first define a parameter set with sequence length S = 256, embedding size E = 60, and $dim_{hid}$ = 16. Considering one decoder block with four time steps, SMHA and MLP, we arrive at a total of 114.24 million SOPs, which translates roughly to \textbf{102.82 $\mu$J}. The consumption can be further reduced to \textbf{53.14$\mu$J} (59.04 million SOPs) by employing only two time steps. Now, applying the same considerations to floating-point operations in the non-spiking decoder block, we arrive at roughly 167.71 million FLOPs or \textbf{768.66$\mu$J}.
These values exclude scaling for both variants, since it is an optimization and not inherently needed. Also, the softmax and GELU activations in the non-spiking decoder block are disregarded because of their complex behavior. Nevertheless, this evidently shows SpikeDecoder's capability to reduce the total power consumption by 86.7\% or 92.3\% compared the non-spiking decoder block, respectively. Even with additional time steps and an increased embedding dimension, SpikeDecoder maintains a highly efficient operation mode that is only a fraction of that in the common ANN variant.

\section{Discussion}
In this work, we aimed to combine the powerful decoder-only Transformer architecture with Spiking Neural Networks. To enable a functional end product, we have focused on practical applications of the corresponding components and deferred some SNN-specific issues or opportunities, such as sophisticated spike encoding and custom backpropagation, to future work. This enabled us to focus more closely on the transition process and the effects of employing spiking neurons for specific model parts. Also, we identified and investigated key structures in the final fully SNN-based decoder model. In addition to the insights we have gained during the conversion from a non-spiking to a spiking architecture, we have implemented an interactive tool to directly compare generated outputs of a common pre-trained ANN decoder, a partially spiking decoder, and the final SpikeDecoder variant.
Finally, we have provided an overview of the full model development and the insights gained through the transition process.

\subsection{Methodology and Findings Review} \label{Review and Conclusion}
Throughout this work, we have analyzed the current research landscape of Transformer-based SNN models, particularly in natural language applications. We went over prerequisites and specific implementation requirements for creating and processing data with spikes. Similarly, we have revisited a number of structural choices made in previous models and evaluated their suitability for our purposes. While we were able to directly adapt some techniques, such as Spiking-Self-Attention, we had to develop custom alternatives for others, such as normalization and time-step unification. We also addressed some implementation preliminaries related to the shift from computer vision to natural language processing and from encoder to decoder. After revisiting a common GPT ANN model, we presented our novel purely spike-based decoder-only architecture, SpikeDecoder, for application to natural language. To illustrate the transition from a non-spiking to a spiking architecture, we have structured the final architecture into four segments and described the individual effects of making them SNN-based and how to compensate for the corresponding performance loss. Additionally, we have re-evaluated several architectural choices, including residual connections and parameterized LIFs. After showing objective performance measurements, we derived the reduced theoretical energy consumption. Our final model achieves stable accuracy on smaller test sets and has the potential to serve as a starting point for future development of transformer-based, directly trained SNN models. A few possibilities to start this development will be presented in the next section.

\subsection{Future Research Opportunities} \label{Future Research Opportunities}
There have already been some indications of potential investigation points mentioned throughout this work. Namely, due to difficulties with GPU parallelization and hardware limitations, we have yet to apply the model on large datasets and explore parameter sizes beyond $\sim$20 million. For this purpose, we have already distributed the text8 dataset in the model repository, which is a lowercase semi-clean version of enwik8, a text corpus made up of Wikipedia entries. To further motivate this direction, we compare generated text excerpts from models trained on a one-million-character dataset for ten epochs in Appendix Figure \ref{fig:TextComparison}, showing considerable improvement in quality. Another issue that could be further investigated is the adaptation of search algorithms without having to employ softmax. In a similar manner, SpikeDecoder has been developed with a character-level application in mind, but it would also be suitable to be applied to whole words, with some changes to the embedding block. We also motivate this approach through Figure \ref{fig:GeneratedWords} in the appendix and include it in the current implementation. Also, regarding the embedding block, the positional encoding used has not been designed with binary structures in mind. In the future, it would be sensible to introduce a more sophisticated strategy that is designed for binary representation. Besides application data, the spike-encoding strategy may also be worth investigating further. We have already briefly mentioned possible alternatives when introducing our use of static repetition for synthetic time steps, which can be incorporated with some minor changes to the embedding block. Since information from earlier processing steps can influence later computations, resembling recurrent dynamics, conducting a more thorough comparison between transformer-based and RNN-based models, such as SpikeGPT, would be very interesting as well.

With regard to the SpikeDecoder architecture, one could further investigate other suitable normalization and backpropagation techniques. Also, there may be alternatives to the presented residual structures that do not degrade performance as much when operating on spike values. As an alternative to the tested dynamic threshold, one could enable other learnable properties, such as the reset value. This may also be a good opportunity to introduce a proper refractory period, which would increase the biological validity of the SNN nodes. Generally, the potential of the architecture has not yet been fully explored, and SpikeDecoder can be seen as a functional proof for the capabilities of a directly trained spiking decoder model and, correspondingly, a starting point for future optimization.

\section*{Conflict of Interest Statement}
%All financial, commercial or other relationships that might be perceived by the academic community as representing a potential conflict of interest must be disclosed. If no such relationship exists, authors will be asked to confirm the following statement: 
The authors declare that the research was conducted in the absence of any commercial or financial relationships that could be construed as a potential conflict of interest.

\section*{Author Contributions}
CB: Conceptualization, Methodology, Visualization, Software and Writing - original draft. FW: Conceptualization, Supervision, Writing - review and editing. AK: Supervision.

\section*{Funding}
The author(s) received no financial support for the research, authorship, and/or publication of this article.

\section*{Data Availability Statement}
The repository for this study can be found on \href{https://github.com/ClaasBeger/SpikeDecoder}{github}.

% ------------------ Bibliography ------------------
% Make sure the file name matches your .bib (assumed bibliography.bib)
\bibliography{bibliography}

@incollection{ghosh2009third,
	title={Third generation neural networks: Spiking neural networks},
	author={Ghosh-Dastidar, Samanwoy and Adeli, Hojjat},
	booktitle={Advances in computational intelligence},
	pages={167--178},
	year={2009},
	publisher={Springer}
}

@Article{maass1997networks,
	title={Networks of spiking neurons: the third generation of neural network models},
	author={Maass, Wolfgang},
	journal={Neural networks},
	volume={10},
	number={9},
	pages={1659--1671},
	year={1997},
	publisher={Elsevier}
}

@article{walter2015nmc,
title = {Neuromorphic implementations of neurobiological learning algorithms for spiking neural networks},
journal = {Neural Networks},
volume = {72},
pages = {152-167},
year = {2015},
note = {Neurobiologically Inspired Robotics: Enhanced Autonomy through Neuromorphic Cognition},
issn = {0893-6080},
doi = {https://doi.org/10.1016/j.neunet.2015.07.004},
author = {Florian Walter and Florian Röhrbein and Alois Knoll}
}

@misc{vaswani2017attention,
	title={Attention Is All You Need}, 
	author={Ashish Vaswani and Noam Shazeer and Niki Parmar and Jakob Uszkoreit and Llion Jones and Aidan N. Gomez and Lukasz Kaiser and Illia Polosukhin},
	year={2017},
	eprint={1706.03762},
	archivePrefix={arXiv},
	primaryClass={cs.CL}
}

@INPROCEEDINGS{9664146,
	author={Mueller, Etienne and Studenyak, Viktor and Auge, Daniel and Knoll, Alois},
	booktitle={2021 7th International Conference on Systems and Informatics (ICSAI)}, 
	title={Spiking Transformer Networks: A Rate Coded Approach for Processing Sequential Data}, 
	year={2021},
	volume={},
	number={},
	pages={1-5},
	doi={10.1109/ICSAI53574.2021.9664146}}

@inproceedings{NEURIPS2021_afe43465,
	author = {Fang, Wei and Yu, Zhaofei and Chen, Yanqi and Huang, Tiejun and Masquelier, Timoth\'{e}e and Tian, Yonghong},
	booktitle = {Advances in Neural Information Processing Systems},
	editor = {M. Ranzato and A. Beygelzimer and Y. Dauphin and P.S. Liang and J. Wortman Vaughan},
	pages = {21056--21069},
	publisher = {Curran Associates, Inc.},
	title = {Deep Residual Learning in Spiking Neural Networks},
	url = {https://proceedings.neurips.cc/paper_files/paper/2021/file/afe434653a898d-a20044041262b3ac74-Paper.pdf},
	volume = {34},
	year = {2021}
}

@article{zhou2022spikformer,
	title={Spikformer: When spiking neural network meets transformer},
	author={Zhou, Zhaokun and Zhu, Yuesheng and He, Chao and Wang, Yaowei and Yan, Shuicheng and Tian, Yonghong and Yuan, Li},
	journal={arXiv preprint arXiv:2209.-15425},
	year={2022}
}

@article{li2022spikeformer,
	title={Spikeformer: A Novel Architecture for Training High-Performance Low-Latency Spiking Neural Network},
	author={Li, Yudong and Lei, Yunlin and Yang, Xu},
	journal={arXiv preprint -arXiv:2211.10686},
	year={2022}
}

@misc{zhou2023spikingformer,
	title={Spikingformer: Spike-driven Residual Learning for Transformer-based Spiking Neural Network}, 
	author={Chenlin Zhou and Liutao Yu and Zhaokun Zhou and Han Zhang and Zhengyu Ma and Huihui Zhou and Yonghong Tian},
	year={2023},
	eprint={2304.11954},
	archivePrefix={arXiv},
	primaryClass={cs.NE}
}

@misc{zhu2023spikegpt,
	title={SpikeGPT: Generative Pre-trained Language Model with Spiking Neural Networks}, 
	author={Rui-Jie Zhu and Qihang Zhao and Jason K. Eshraghian},
	year={2023},
	eprint={2302.13939},
	archivePrefix={arXiv},
	primaryClass={cs.CL}
}

@misc{SpikingJelly,
	title = {SpikingJelly},
	author = {Fang, Wei and Chen, Yanqi and Ding, Jianhao and Chen, Ding and Yu, Zhaofei and Zhou, Huihui and Timothée Masquelier and Tian, Yonghong and other contributors},
	year = {2020},
	howpublished = {\url{https://github.com/fangwei123456/spikingjelly}},
	note = {[Accessed 25-Jun-2023]},
}

@misc{Lenz_2023,
	title={SNN library benchmarks}, url={https://open-neuromorphic.org/p/snn-library-benchmarks/}, journal={Open Neuromorphic}, author={Lenz, Gregor}, year={2023}, month={8}}

@article{wang2022understanding,
	title={Understanding the Failure of Batch Normalization for Transformers in NLP},
	author={Wang, Jiaxi and Wu, Ji and Huang, Lei},
	journal={Advances in Neural Information Processing Systems},
	volume={35},
	pages={37617--37630},
	year={2022}
}

@misc{shen2020powernorm,
	title={PowerNorm: Rethinking Batch Normalization in Transformers}, 
	author={Sheng Shen and Zhewei Yao and Amir Gholami and Michael W. Mahoney and Kurt Keutzer},
	year={2020},
	eprint={2003.07845},
	archivePrefix={arXiv},
	primaryClass={cs.CL}
}

@misc{Karpathy,
	title={Karpathy/minGPT: A minimal pytorch re-implementation of the openai GPT (generative pretrained transformer) training}, url={https://github.com/karpathy/minGPT/tree/master}, journal={GitHub}, year = {2022}, author={Karpathy, Andrej}
}

@INPROCEEDINGS{6757323,
	author={Horowitz, Mark},
	booktitle={2014 IEEE International Solid-State Circuits Conference Digest of Technical Papers (ISSCC)}, 
	title={1.1 Computing's energy problem (and what we can do about it)}, 
	year={2014},
	volume={},
	number={},
	pages={10-14},
	doi={10.1109/ISSCC.2014.6757323}}

@misc{neftci2019surrogate,
	title={Surrogate Gradient Learning in Spiking Neural Networks}, 
	author={Emre O. Neftci and Hesham Mostafa and Friedemann Zenke},
	year={2019},
	eprint={1901.09948},
	archivePrefix={arXiv},
	primaryClass={cs.NE}
}

\newpage

\appendix

\section{Additional Figures and Examples}

\begin{figure}[htb]%
	\centering%
	%
	% Including .png
	\includegraphics[width=80mm]{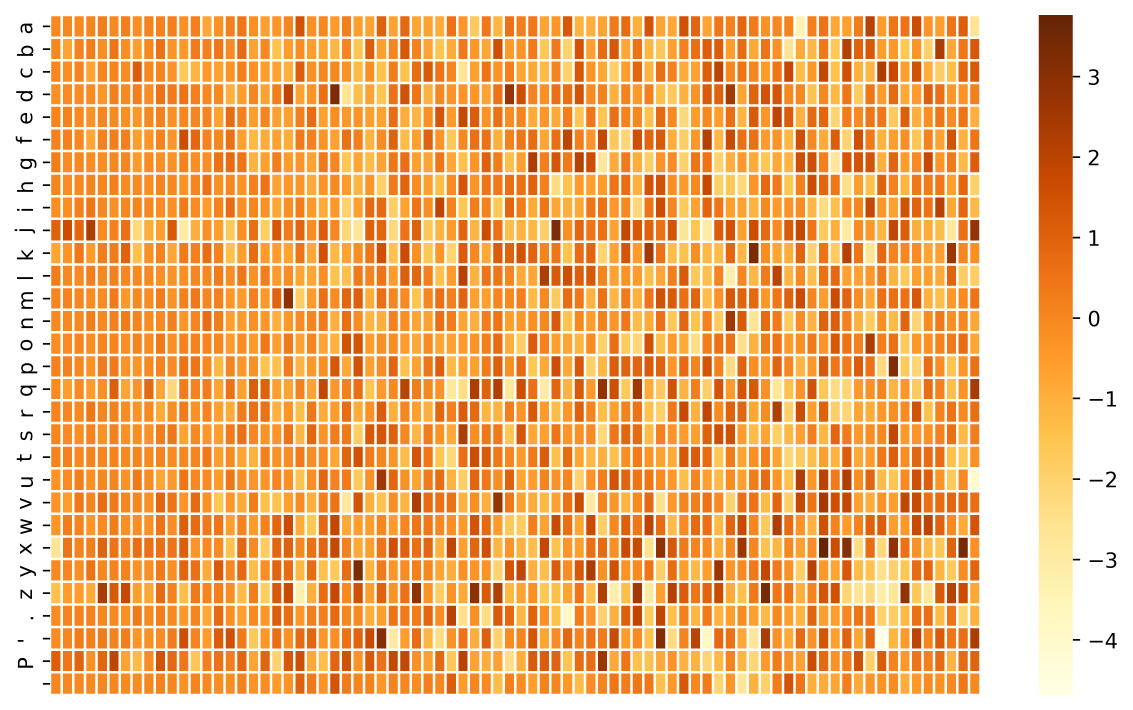}%
	\includegraphics[width=80mm]{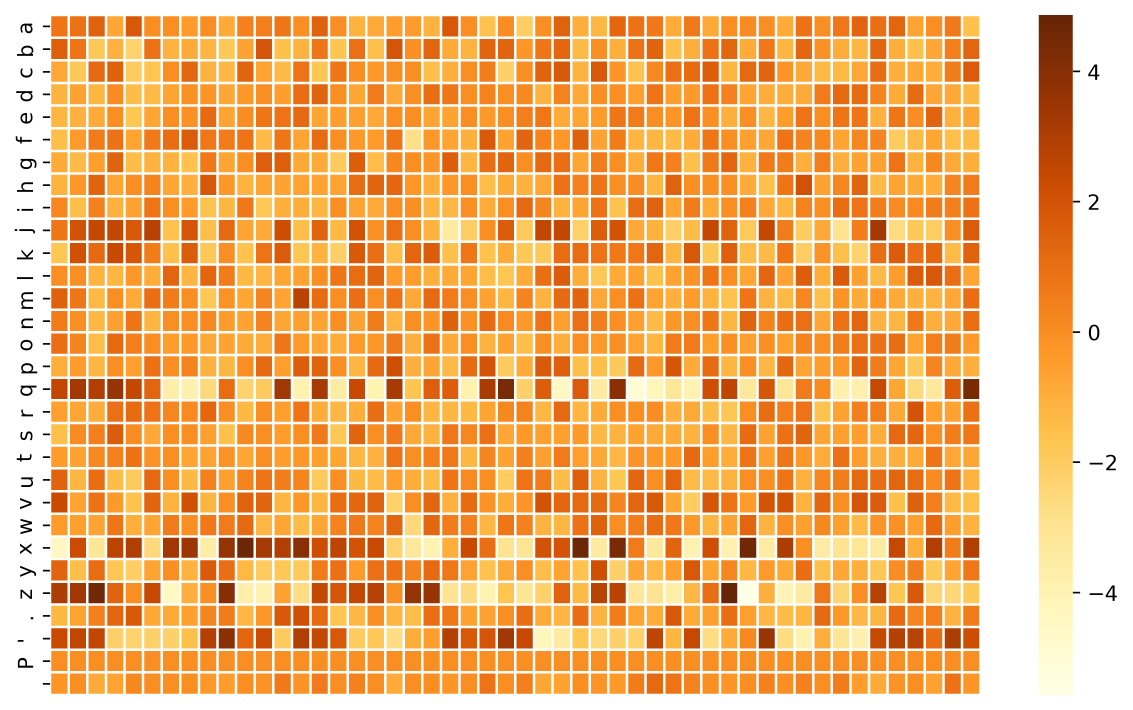}%
	\vspace{0.1cm}
	\includegraphics[width=80mm]{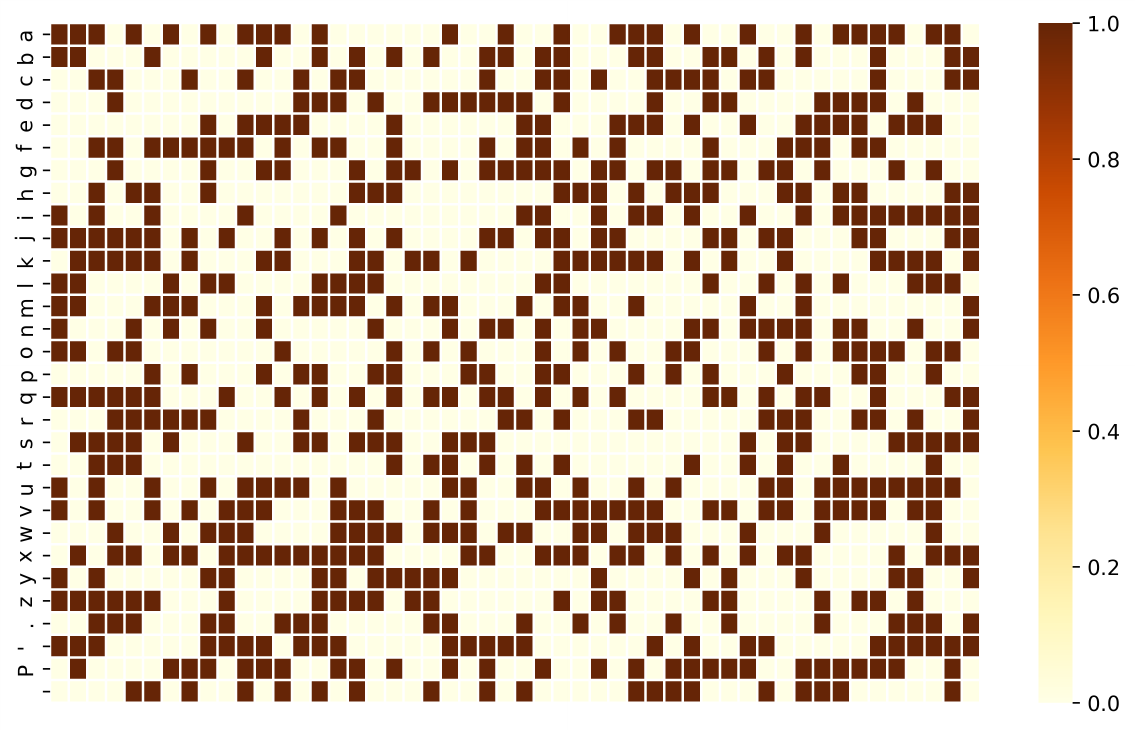}%
	\caption{The different token embedding vectors, with corresponding characters on the y-axis. The first table shows the embedding vector for the non-spiking decoder model, which has a dimension of 80. The second table shows the raw values of the embedding object from the Partially Spiking decoder model with spike degree 1, which has a reduced dimension of 50 (the rest is used for positional encoding). The third table shows the true token embedding values, which are calculated using the Heaviside function.}%
	\label{fig:Token_Embeddings}%
\end{figure}%
\begin{figure}[htb]%
	\begin{center}
		\includegraphics[width=\textwidth]{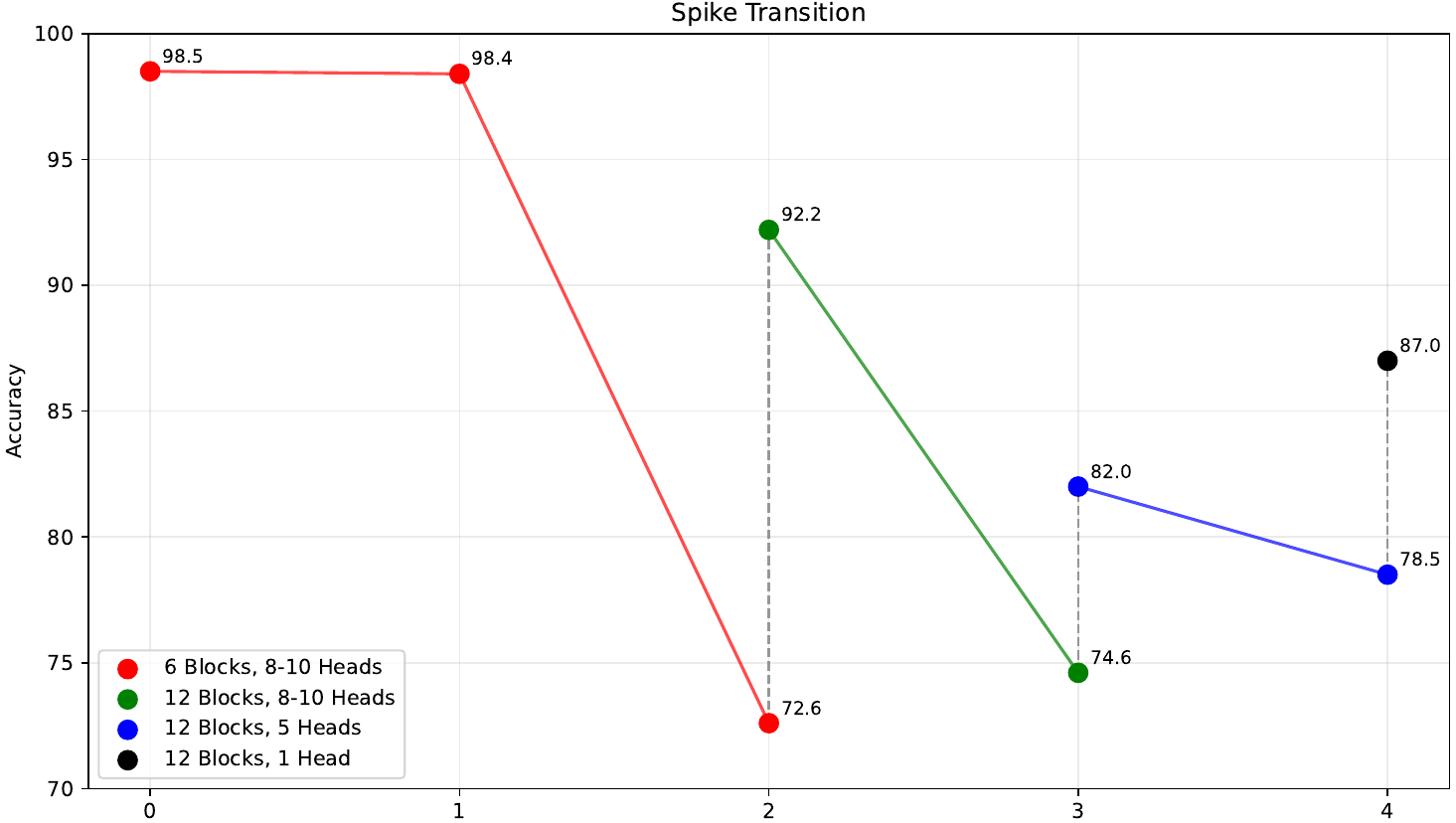}
		\caption{Display of the accuracy values by spike degree. After large drops, we modify the model through adding depth or reducing the number of heads, in order to stay as close as possible to the original performance.}%
		\label{fig:Transition Graph}%
	\end{center}
\end{figure}%

\begin{figure}[htb] % use [htbp] instead if you prefer normal floating
  \centering

  \begin{subfigure}{\textwidth}
    \centering
    \includegraphics[width=0.95\linewidth]{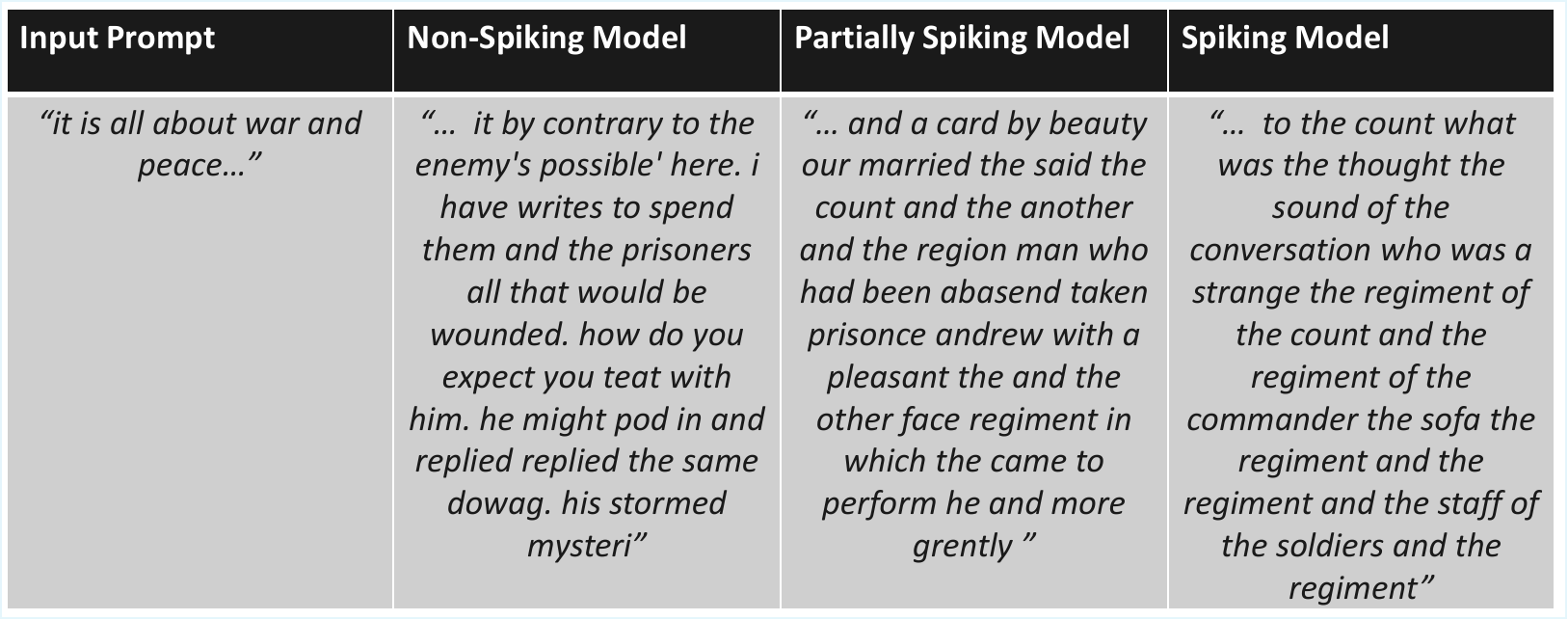}
    \caption{Comparison of generated text excerpts from the three main model types trained on one million characters. Partially spiking refers to spike degree 2 in this context.}
    \label{fig:TextComparison}
  \end{subfigure}

  \vspace{1em}

  \begin{subfigure}{\textwidth}
    \centering
    \includegraphics[width=0.95\linewidth]{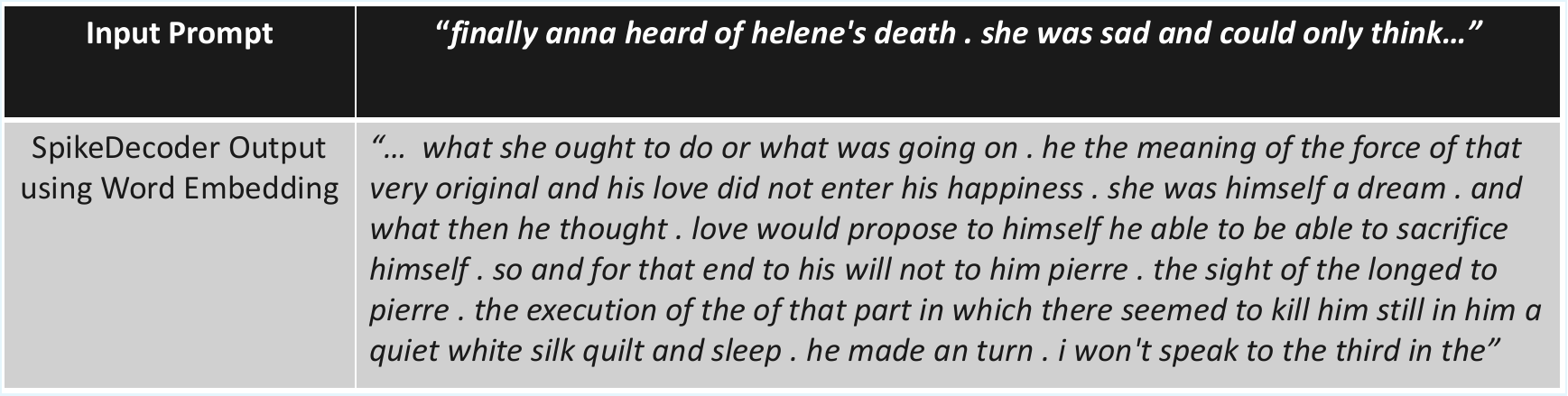}
    \caption{SpikeDecoder output using a full word embedding. The model noteably infers the name of Pierre, Helenes widower, but ignores Annas inclusion in the prompt.}
    \label{fig:GeneratedWords}
  \end{subfigure}

  \caption{Generated text analyses.}
  \label{fig: generated_subfigures}
\end{figure}

\end{document}